# Probabilistic Hybrid Action Models
# for Predicting Concurrent Percept-driven Robot Behavior

**Michael Beetz**                                                              BEETZ@IN.TUM.DE
*Department of Computer Science IX, Technische Universität München,*
*Boltzmannstr. 3, D-81667 Garching, Germany,*
**Henrik Grosskreutz**                                          GROSSKREUTZ@CS.RWTH-AACHEN.DE
*Department of Computer Science, Aachen University of Technology*
*D-52056 Aachen, Germany*

## Abstract

This article develops **Probabilistic Hybrid Action Models (PHAMs)**, a realistic causal model for predicting the behavior generated by modern percept-driven robot plans. PHAMs represent aspects of robot behavior that cannot be represented by most action models used in AI planning: the temporal structure of continuous control processes, their non-deterministic effects, several modes of their interferences, and the achievement of triggering conditions in closed-loop robot plans.

The main contributions of this article are: (1) PHAMs, a model of concurrent percept-driven behavior, its formalization, and proofs that the model generates probably, qualitatively accurate predictions; and (2) a resource-efficient inference method for PHAMs based on sampling projections from probabilistic action models and state descriptions. We show how PHAMs can be applied to planning the course of action of an autonomous robot office courier based on analytical and experimental results.

## 1. Introduction

Most autonomous robots are equipped with restricted, unreliable, and inaccurate sensors and effectors and operate in complex and dynamic environments. A successful approach to deal with the resulting uncertainty is the use of controllers that prescribe the robots' behavior in terms of *concurrent reactive plans (CRPs)* — plans that specify how the robots are to react to sensory input in order to accomplish their jobs reliably (e.g., McDermott, 1992a; Beetz, 1999). Reactive plans are successfully used to produce situation specific behavior, to detect problems and recover from them automatically, and to recognize and exploit opportunities (Beetz et al., 2001). These kinds of behaviors are particularly important for autonomous robots that have only uncertain information about the world, act in dynamically changing environments, and are to accomplish complex tasks efficiently.

Besides reliability and flexibility, foresight is another important capability of competent autonomous robots (McDermott, 1992a). Temporal projection, the computational process of predicting what will happen when a robot executes its plan, is essential for the robots to plan their intended courses of action successfully. To be able to project their plans, robots must have causal models that represent the effects of their actions. Most robot action planners use representations that include discrete action models and plans that define partial orders on actions. Therefore, they cannot automatically generate, reason about, and revise modern reactive plans. This has two important drawbacks. First, the planners cannot accurately predict and diagnose the behavior generated by their plans because they abstract away from important aspects of reactive plans. Second, the plan-ners cannot exploit the control structures provided by reactive plan languages to make plans more flexible and reliable.

In this article we develop **PHAM**s (Probabilistic Hybrid Action Models), action models that have the expressiveness for the accurate prediction of behavior generated by concurrent reactive plans. To the best of our knowledge, PHAMs are the only action representation used in action planning that provides programmers with means for describing the interference of simultaneous, concurrent effects, probabilistic state and action models, as well as exogenous events. PHAMs have been successfully applied by an autonomous robot office courier and a museum tour-guide robot to make predictions of full-size robot plans during the execution of these plans (Beetz, 2001).

This article makes several important contributions to the area of decision-theoretic robot action planning. First, we describe PHAMs, formal action models that allow for the prediction of the qualitative behavior generated by concurrent reactive plans. Second, we show how PHAMs can be implemented in a resource efficient way such that predictions based on PHAMs can be performed by robots while executing their plans. Third, we apply the plan projection method to probabilistic prediction-based schedule debugging and analyze it in the context of a robot office courier (Beetz, 2001).

Before starting with the technical part of the article we would like to make several remarks. In this article we restrict ourselves to navigation actions and model them exactly as they are implemented in one of the most successful autonomous robot navigation systems (Burgard et al., 2000). The reason is that we want to close the gap between action models used in AI planning systems and the control programs that are used by autonomous robots and the behavior they produce. The control programs that we model have proven themselves to achieve reliable, high performance navigation behavior. In the Minerva experiment, they controlled the navigation in a crowded museum for more than 93 hours. During their execution, the navigation plans have been revised by a planning module about 3200 times without causing any deadlocks between interacting, concurrent control processes (Beetz, 2002a; Beetz et al., 2001). In robot office courier experiments, we have applied plan revision methods that enabled the robot to plan ahead for about 15-25 minutes. We consider this to be a time scale sufficient for improving the robot's performance through planning. However, the performance gains that can in principle be achieved through navigation planning are often small compared to those that can be achieved by planning manipulation tasks.

Although we use navigation as our only example, the same modeling techniques apply to other mechanisms of autonomous robots, such as vision (Beetz et al., 1998), communication (Beetz & Peters, 1998), and manipulation (Beetz, 2000) equally well. The reasons that we do not cover these kinds of actions in this article are that they require additional reasoning capabilities and at the moment these models can only be validated with respect to robot simulations. The additional robot capabilities that would have to be modeled include symbol grounding/object recognition (Beetz, 2000), changing states of objects, and more thorough models of the belief states of robots (Schmitt et al., 2002). Addressing these issues is well beyond the scope of this article.

In the remainder of the article we introduce the basic conceptualization underlying PHAMs and describe two realizations of them: one for studying their formal properties and another one targeted at their efficient implementation. We also show how PHAMs are employed in the context of transformational robot planning.

The article is organized as follows. Section 2 describes everyday activity as our primary class of application problems. We introduce concurrent reactive plans (CRPs) as means for producing characteristic patterns of everyday activity and identify technical problems in the prediction of the physical robot behavior that CRPs generate. Section 3 explains how the execution of CRPs and the physical and computational effects of plan execution can be modeled using PHAMs. PHAMs describe the behavior of the robot as a sequence of control modes where in each mode the continuous behavior is specified by a control law. Mode transitions are triggered by the controlled system satisfying specified mode transition conditions. We then introduce a set of predicates that we use to represent our conceptualization formally. Section 4 and 5 describe two different approaches to predicting the behavior produced by concurrent reactive plans in the context of PHAMs. In the first one the behavior is approximated by discretizing time into a sequence of clock ticks that can be made arbitrarily dense. This model is used to derive formal properties for the projection of concurrent reactive plans. The second approach, described in Section 5, describes a much more efficient approach to the projection of CRPs. In this approach only those time ticks are explicitly considered and represented where discrete events may occur. At all other time instances the system state can be inferred through interpolation using the control laws of the respective modes. This is the projection mechanism that is used at execution time on board the robots. We show how this implementation of PHAMs is employed for prediction-based tour scheduling for an autonomous robot office courier. We conclude with an evaluation and a discussion of related work.

## 2. Structured Reactive Controllers and the Projection of Delivery Tour Plans

Plan-based robot control has been successfully applied to tasks such as the control of space probes (Muscettola et al., 1998b), disaster management and surveillance (Doherty et al., 2000), and the control of mobile robots for office delivery (Simmons et al., 1997; Beetz et al., 2001) and tourguide scenarios (Alami et al., 2000; Thrun et al., 2000). A class of tasks that has received little attention is the plan-based robot control for everyday activity in human living and working environments, tasks that people are usually very good at.

To get a better intuition of the activity patterns to be produced in everyday activity, let us consider the chores of a hypothetical household robot. Household chores entail complex routine jobs such as cooking dinner, cleaning the kitchen, loading the dish washer, etc. The routine jobs are typically performed in parallel. A household robot might have to clean up the living room while the soup is cooking on the stove. While cleaning up, the phone might ring and the robot has to interrupt cleaning in order to go and answer the phone. After having completed the telephone call the robot has to continue cleaning right where it stopped. Thus, the robot's activity must be concurrent, percept-driven, interruptible, flexible, and robust, and it requires foresight.

The fact that people manage and execute their daily tasks effectively suggests, in our view, that the nature of everyday activity should permit agents to make assumptions that simplify the computational tasks required for competent activity. As Horswill (1996) puts it, everyday life must provide us with some loopholes, structures and constraints that make activity tractable.

We believe that in many applications of robotic agents that are to perform everyday activities, the following assumptions are valid and allow us to simplify the computational problems of controlling a robot competently:

1. Robotic agents are familiar with the activities for satisfying individual tasks and the situations that typically occur while performing them. They carry out everyday activities over and over again and are confronted with the same kinds of situations many times. As a consequence, conducting individual everyday activities can be learned from experience and is simple in the sense that it does not require a lot of plan generation from first principles.

2. Appropriate plans for satisfying multiple, possibly interfering, tasks can be determined in a greedy manner. The robot can first determine a default plan performing the individual tasks concurrently with some additional ordering constraints through simple and fast heuristic plan combination methods. The robot can then avoid the remaining interferences between its sub-activities by predicting and forestalling them.

3. Robotic agents can monitor the execution of their activities and thereby detect situations in which their intended course of action might fail to produce the desired effects. If such situations are detected, the robots can adapt their intended course of action to the specific situations they encounter, if necessary based on foresight.

In our previous research we have proposed Structured Reactive Controllers (SRCs) as a computational model for the plan-based control of everyday activity. SRCs are collections of concurrent reactive control routines that adapt themselves to changing circumstances during their execution by means of planning. SRCs are based upon the following computational principles:

1. SRCs are equipped with a library of plan schemata for routine tasks in common situations. These plan schemata are — for now — provided by programmers and designed to have high expected utility at the cost of not having to deal with all conceivable problems. We know from our AI courses that plans that check the tailpipes every time before starting a car have typically lower expected utility than the ones that do not check them, even though having no bananas stuck in the tailpipe is a necessary precondition for starting a car successfully.

   The robustness, flexibility, and reactivity of plan schemata is achieved by implementing them as *concurrent percept-driven* plans — even at the highest level of abstraction. The plans employ control structures including conditionals, loops, program variables, processes, and subroutines. They also make use of high-level constructs (interrupts, monitors) to synchronize parallel actions and make plans reactive and robust by incorporating sensing and monitoring actions and reactions triggered by observed events. Goals of sub-plans are represented explicitly as annotations such that planning algorithms can infer the purpose of sub-plans automatically.

2. SRCs have fast heuristic methods for putting plans together from routine activities. They are able to predict problems that are likely to occur and revise their course of action to avoid them. Predictive plan debugging requires the SRC to reason through, and predict the effects of, highly conditional and flexible plans — the subject of this article.

3. SRCs perform execution time plan management. They run processes that monitor the beliefs of the robot and are triggered by certain belief changes. These processes revise plans while they are executed.

Structured reactive controllers work as follows. When given a set of requests, structured reactive controllers retrieve routine plans for individual requests and execute the plans concurrently. These routine plans are general and flexible — they work for standard situations and when executed concurrently with other routine plans. Routine plans can cope well with partly unknown and changing environments, run concurrently, handle interrupts, and control robots without assistance over extended periods. For standard situations, the execution of these routine plans causes the robot to exhibit an appropriate behavior in achieving their purpose. While they execute routine plans, the robot controllers also try to determine whether their routines might interfere with each other and watch out for exceptional situations. If they encounter exceptional situations they will try to anticipate and forestall behavior flaws by predicting how their routine plans might work in this kind of situation. If necessary, they revise their routines to make them robust for the respective kinds of situations. Finally, they integrate the proposed revisions smoothly into their ongoing course of actions.

## 2.1 Plan-based Control for a Robot Office Courier

Before we describe our approach to predicting concurrent percept-driven robot behavior we first give a comprehensive example of a plan-based robot office courier performing a delivery tour and exhibiting aspects of everyday activity. The description of the example includes the presentation of key plan schemata used by the robot, a sketch of the heuristic plan combination method, the prediction of behavior flaws, and the revision of delivery plans. This example run has been performed with the mobile robot RHINO acting as a robot office courier (Beetz, 2001; Beetz, Bennewitz, & Grosskreutz, 1999).

### 2.1.1 PLANS AND PLAN SCHEMATA OF THE ROBOT COURIER

The robot courier is equipped with a library of plan schemata for its standard tasks including delivering items, picking up items, and navigating from one place to another. The presentation of the plans and plan schemata proceeds bottom up. We start with the low-level plans for navigation and end with the comprehensive object delivery plans.

A low-level navigation plan specifies how the robot is to navigate from one location in its environment, typically its current position, to another one, its destination. Figure 1 depicts such a low-level navigation plan for going from a location in room A-117 to the location 5 in room A-111. The plan consists of two components: a sequence of intermediate target points (the locations indexed by the numbers 1 to 5 in Figure 1) to be sequentially visited by the robot and a specification of when and how the robot is to adapt its travel modes as it follows the navigation path. In many environments it is advantageous to adapt the travel mode to the surroundings: to drive carefully (and therefore slowly) within offices because offices are cluttered, to switch off the sonars when driving through doorways (to avoid sonar crosstalk), and to drive quickly in the hallways. The second part of the plan is depicted through regions with different textures for the different travel modes "office", "hallway," and "doorway." Whenever the robot crosses the boundaries between regions it adapts the parameterization of the navigation system. Thus, low-level navigation plans start and terminate navigation processes and change the parameterization of the navigation system through control mode switches (SET-NAVIGATION-MODE) and adding and deleting intermediate target points (MOVE-TO).

We specify reactive plans in RPL (McDermott, 1991; Beetz & McDermott, 1992), a plan language that provides high-level control structures for specifying concurrent, event-driven robot behavior. The pseudo code in Figure 2 sketches the initial part of the plan depicted in Figure 1. The plan for leaving an office consists of two concurrent sub-plans: one for following the (initial part of the) prescribed path and one for adapting the travel mode. The second sub-plan adapts the navigation mode of the robot dynamically. Initially, the navigation mode is set to "office". Upon entering and leaving the doorway the navigation mode is adapted. The plan uses *fluents*, conditions that are updated asynchronously based on new sensor readings. Fluents can trigger (*whenever*) and terminate (*wait for*) plan steps.

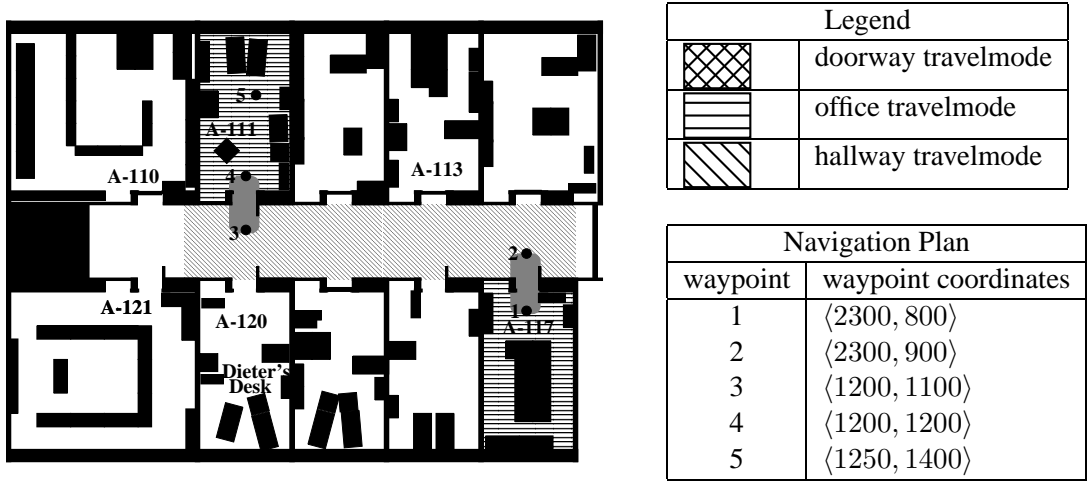

Figure 1: Graphical representation of a navigation plan. Topological navigation plan for navigating from room A-117 to A-111 with regions indicating different travel modes and small black circles indicating additional navigation path constraints.

*execute concurrently*
    *execute-in-order*
        MOVE-TO (*1*); *wait for* (***go-to-completed?***);
        MOVE-TO (*2*); *wait for* (***go-to-completed?***);
    *with local fluents* ***distance-to-doorway***
                $\leftarrow$ ***fluent-network*** ($|\ \langle x, y \rangle - \langle x_{dw}, y_{dw} \rangle\ |$)
            ***entering-dw?-fl*** $\leftarrow$ ***distance-to-doorway*** $< 1$m
            ***entering-hw?-fl*** $\leftarrow$ ***distance-to-doorway*** $> 1$m
        *execute-in-order*
            SET-NAVIGATION-MODE(office); *wait for* (***entering-dw?-fl***);
            SET-NAVIGATION-MODE(doorway); *wait for* (***entering-hw?-fl***);
            SET-NAVIGATION-MODE(hallway)

Figure 2: The plan sketches the specification of a navigation process for leaving an office. The two components following the prescribed path and adapting the travel mode are implemented as concurrent sub-plans. The second component uses a fluent to measure the distance to the center of the doorway and two dependent fluents that signal the robot's entering and leaving the doorway. Initially, the travel mode is set to "office". Upon entering and leaving the doorway the travel mode is adapted.

The low-level navigation plan instances are used by higher-level navigation plans that make the navigation processes flexible, robust, and embeddable into concurrent task contexts. The higher-level plans generate the low-level plans based on the robot's map of its environment (GENERATE-NAV-PLAN). A slightly simplified version of this high-level plan is listed below.

```
highlevel-plan ACHIEVE(loc(rhino, ⟨x, y⟩))
   1    with cleanup routine ABORT-NAVIGATION-PROCESS
   2    do with valve wheels
   3       do loop
   4              try in parallel
   5              wait for navigation-interrupted?
   6              with local vars NAV-PLAN ← GENERATE-NAV-PLAN(c,d)
   7              do swap-plan (NAV-PLAN,NAV-STEP)
   8                 named subplan NAV-STEP
   9                 do DUMMY
  10           until IS-CLOSE?(⟨x, y⟩)
```

We explain the plan going from the inner parts, which generate the robot behavior, to the outer ones, which modify the behavior. Lines 6 to 8 make the navigation plan independent of its starting position and thereby more general: given a destination $d$, the plan piece computes a low-level navigation plan from the robot's current location $c$ to $d$ using the map of the environment and executes it (Beetz & McDermott, 1996).

In order to run navigation plans in less constrained task contexts we must prevent other — concurrent — routines from directing the robot to different locations while the navigation plan is executed. We accomplish this by using semaphores or "valves", which can be requested and released. Any plan asking the robot to move or stand still must request the valve *wheels*, perform its actions only after it has received *wheels*, and release *wheels* after it is done. This is accomplished by the statement <u>with valve</u> in line 2.

In many cases processes with higher priorities must move the robot urgently. In this case, blocked valves are simply pre-empted. To make our plan *interruptible*, robust against such interrupts, the plan has to do two things. First, it has to detect when it gets interrupted and second, it has to handle such interrupts appropriately. This is done by a loop that generates and executes navigation plans for the navigation task until the robot is at its destination. We make the routine cognizant of interrupts by using the fluent **navigation-interrupted?**. Interrupts are handled by terminating the current iteration of the loop and starting the next iteration, in which a new navigation plan starting from the robot's new position is generated and executed. Thus, the lines 3-5 make the plan interruptible.

To make the navigation plan *transparent* we name the routine plan **ACHIEVE**(*loc(rhino,⟨x,y⟩)*) and thereby enable the planning system to infer the purpose of the sub-plan syntactically. Interruptible and embeddable plans can be used in task contexts with higher priority concurrent sub-plans. For instance, a monitoring plan used by our controller estimates the opening angles of doors whenever the robot passes one. Another monitoring plan localizes the robot actively whenever it has lost track of its position.

To facilitate online rescheduling we have modularized the plans with respect to the locations where sub-plans are to be executed using the <u>**at location**</u> plan schema. The <u>**at location**</u> ⟨x,y⟩ p plan schema specifies that plan $p$ is to be performed at location ⟨x,y⟩. Here is a simplified version of the plan schema for <u>**at location**</u>.

> *named subplan* $N_i$
> <u>*do*</u> <u>*at location*</u>    $\langle x, y \rangle\, p$ *by*
>     <u>*with valve*</u> *wheels*
>     <u>*do*</u> <u>*with local vars*</u> DONE? $\leftarrow$ FALSE
>         <u>*do*</u> *loop*
>             *try in parallel*
>             <u>*wait for*</u> ***Task-Interrupted?***$(N_i)$
>             *sequentially*
>             <u>*do*</u> NAVIGATE-TO$\langle x, y \rangle$
>                 $p$
>                 DONE? $\leftarrow$ TRUE
>         <u>*until*</u> DONE? $=$ TRUE

The plan schema accomplishes the performance of plan $p$ at location $\langle$x,y$\rangle$ by navigating to the location $\langle$x,y$\rangle$, performing sub-plan $p$, and signalling that $p$ has been completed (the inner sequence). The <u>*with valve*</u> statement obtains the semaphore *wheels* that must be owned by any process changing the location of the robot. The loop makes the execution of $p$ at $\langle$x,y$\rangle$ robust against interrupts from higher priority processes. Finally, the <u>*named sub-plan*</u> statement gives the sub-plan a symbolic name that can be used for addressing the sub-plan for scheduling purposes and in plan revisions. Using the <u>*at location*</u> plan schema, a plan for delivering an object $o$ from location $p$ to location $d$ can be roughly specified as a plan that carries out *pickup(o)* at location $p$ and *put-down(o)* at location $d$ with the additional constraint that *pickup(o)* is to be carried out before *putdown(o)*. If every sub-plan $p$ that is to be performed at a particular location $l$ has the form <u>*at location*</u> $\langle$x,y$\rangle$ $p$, then a scheduler can traverse the plan recursively and collect the <u>*at location*</u> sub-plans and install additional ordering constraints on these sub-plans to maximize the plan's expected utility.

To allow for smooth integration of revisions into ongoing scheduled activities, we designed the plans such that each sub-plan keeps a record of its execution state and, if started anew, skips those parts of the plan that no longer have to be executed (Beetz & McDermott, 1996). We made the plans for single deliveries restartable by equipping the plan $p$ with a variable storing the execution state of $p$ that is used as a guard to determine whether or not a sub-plan is to be executed. The variable has three possible values: *to-be-acquired* denoting that the object must still be acquired; *loaded* denoting that the object is loaded; and *delivered* denoting that the delivery is completed. The plan schema for the delivery of a single object consists of two fairly independent plan steps: the pick-up and the put-down step.

> <u>*if*</u> EXECUTION-STATE$(p, \textit{to-be-acquired})$
>     <u>*then*</u> AT-LOCATION    L    PICK-UP$(o)$
> <u>*if*</u> EXECUTION-STATE$(p, \textit{loaded})$
>     <u>*then*</u> AT-LOCATION    D    PUT-DOWN$(o)$

### 2.1.2 GENERATING DEFAULT DELIVERY PLANS

The heuristic plan generator for delivery tours is simple: it inserts the pick-up and put-down sub-plans of all delivery requests into the overall plan and determines an appropriate order on the <u>*at location*</u> sub-plans. The ordering is determined by a heuristic that performs a simple topological sort on the sub-plans based on the locations where the sub-plans are to be executed. The heuristic considers additional constraints such as executing pick-up steps always before the respective put-down plan-steps.

### 2.1.3 PREDICTION-BASED PLAN DEBUGGING BY THE ROBOT OFFICE COURIER

Let us now contemplate a specific scenario in which the robot office courier RHINO performs an office delivery that requires the prediction and forestalling of plan failures at execution time. Consider the following situation in the environment pictured in Figure 3. A robot office courier is to deliver a letter in a yellow envelope from room A-111 to A-117 (*cmd-1*) and another letter for which the envelope's color is unknown from A-113 to A-120 (*cmd-2*). The robot has already tried to accomplish *cmd-2* but because it has recognized room A-113 as closed (using its range sensors) it revises its intended course of action into achieving *cmd-2* opportunistically. That is, if it later detects that A-113 is open it will interrupt its current activity and reconsider its intended course of action under the premise that the steps for accomplishing *cmd-2* are executable.

To perform its tasks quickly the robot schedules the pick-up and delivery actions to minimize execution time and assure that letters are picked up before they are delivered. To ensure that the schedules will work, the robot has to take into account how its own state and the world changes as it carries out the scheduled activities. Aspects of states that the robot has to consider when scheduling its activities are the locations of the letters. Constraints on the state variables that schedules have to satisfy are that they only ask the robot to pick up letters that are currently at the robot's location and that the robot does not carry two letters in envelopes with the same color.

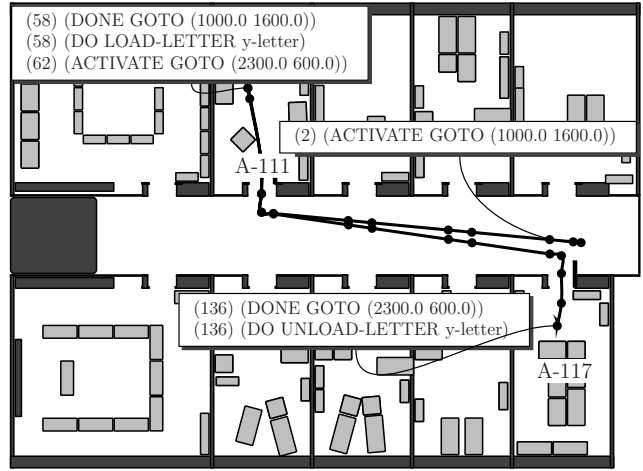

Figure 3: A possible projected execution scenario for the initial plan. The opportunity of loading the letter of the unknown color is ignored.

Suppose our robot is standing in front of room A-117. The belief state of the robot contains probabilities for the colors of letters on the desk in A-113. The robot also has received some evidence that A-113 has been opened in the meantime. Therefore its belief state assigns probability $p$ for the value true of random variable open-A113.

This update of the belief state requires the robot to reevaluate its options for accomplishing its jobs with respect to its changed belief state. Executing its current plan without modifications might yield mix ups because the robot might carry two letters in envelopes with the same color. The different options are: (1) to skip the opportunity, (2) to ask immediately for the letter from A-113 to be put into an envelope that is not yellow (to exclude mix ups when taking the opportunity later); (3) to constrain later parts of the schedule such that no two yellow letters will be carried even when the letter in A-113 turns out to be yellow; and (4) to keep picking up the letter in A-113 as an opportunistic sub-plan. Which option the robot should take depends on its belief state with respect to the states of doors and locations of letters. To find out which schedules will probably work, in particular, which ones might result in mixing up letters, the robot must apply a model of the world dynamics to the state variables.

With respect to this belief state, different scenarios are possible. The first one, in which A-113 is closed, is pictured in Figure 3. Points on the trajectories represent predicted events. The events without labels are actions in which the robot changes its heading (on an approximated trajectory) or events representing sensor updates generated by passive sensing processes. For example, a passive sensor update event is generated when the robot passes a door. In this scenario no intervention by prediction-based debugging is necessary and no flaw is projected.

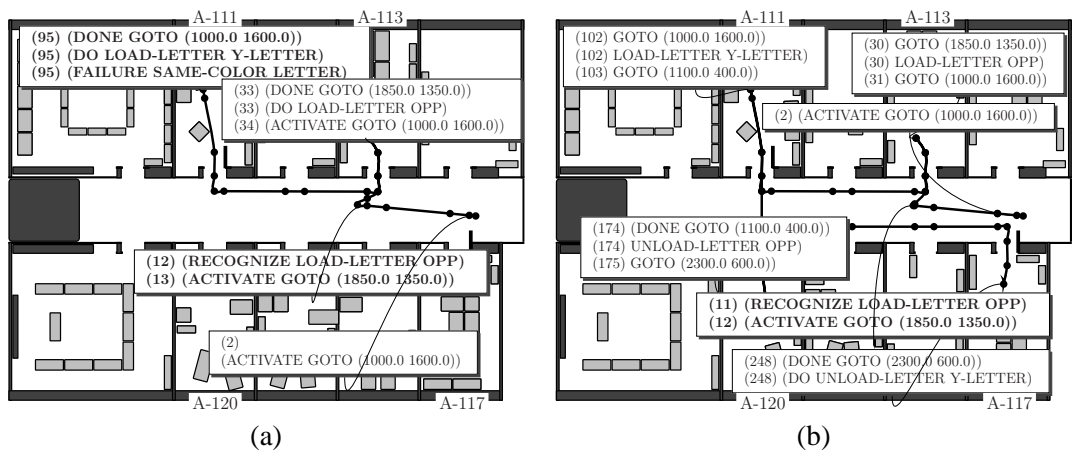

(a)                                                                 (b)

Figure 4: Two possible predicted scenarios for the opportunity being taken. In scenario (a) the letter turns out to have the same color as the one that is to be loaded afterwards. Therefore, the second loading fails. In scenario (b) the letter turns out to have a different color than the one that is to be loaded afterwards. Therefore, the second loading succeeds.

In the scenarios in which office A-113 is open the controller is projected to recognize the opportunity and to reschedule its enabled plan steps as described above.[1] The resulting schedule asks the robot to enter A-113 first, and pickup the letter for *cmd-2*, then enter A-111 and pick up the letter for *cmd-1*, then deliver the letter for *cmd-2* in A-120, and the last one in A-117. This category of scenarios can be further divided into two categories. In the first sub-category shown in Figure 4(a) the letter to be picked up is yellow. Performing the pickup thus would result in the robot carrying two yellow letters and therefore an execution failure is signalled. In the second sub-category shown in Figure 4(b) the letter has a different color and therefore the robot is projected to succeed by taking the same course of action for all these scenarios. Note that the possible flaw is introduced by the reactive rescheduling because the rescheduler does not consider how the state of the robot will change in the course of action, in particular that a state may be caused in which the robot is to carry two letters with the same color.

---

1. Another category of scenarios is characterized by A-113 becoming open after the robot has left A-111. This may also result in an execution failure if the letter loaded in A-113 is yellow, but is not discussed here any further.

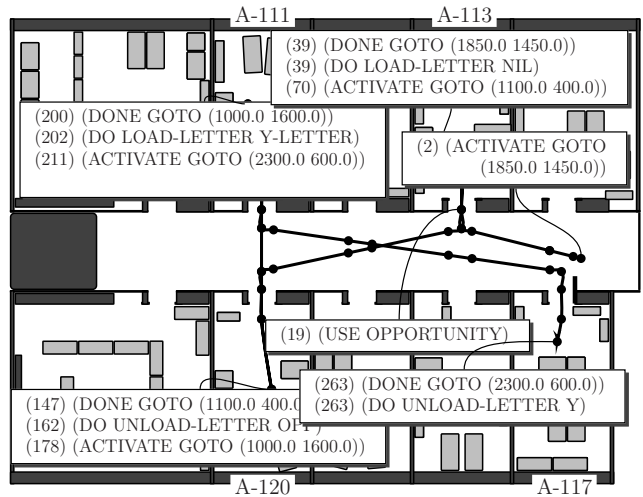

Figure 5: Projected scenario for a plan suggested by the plan debugger. The letter with the unknown color is picked up and also delivered first. This plan is a little less efficient but avoids the risk of not being able to load the second letter.

In this case, the plan-based controller will probably detect the flaw if it is likely with respect to the robot's belief state. This enables the debugger to forestall the flaw, for instance, by introducing an additional ordering constraint, or by sending an email that increases the probability that the letter will be put into a particular envelope. These are the revision rules introduced in the last section. Figure 5 shows a projection of a plan that has been revised by adding the ordering constraint that the letter for A-120 is delivered before entering A-111.

Figure 6(a) shows the event trace generated by the initial plan and *executed* with the RHINO control system (Thrun et al., 1998) for the critical scenario without prediction based schedule debugging; Figure 6(b) shows the one with the debugger adding the additional ordering constraint. This scenario shows that reasoning about the future execution of plans enables the robot to improve its behavior.

In this article, we describe the probabilistic models of reactive robot behavior that are necessary to predict scenarios such as the one described above for the purpose of prediction-based plan debugging.

## 2.2 The Projection of Low-level Navigation Plans

Now that we know what the robot plans look like we can turn to the question of how to predict the effects of executing a delivery plan. The input data for plan projection are the probabilistic beliefs of the robot with respect to the current state of the world, probabilistic models of exogenous events that are assumed to be Poisson distributed, probabilistic models of low-level plans, and probabilistic rules for guessing missing pieces of information. The output of the projection process is a sequence of dated events along with the estimated state at the time of each event.

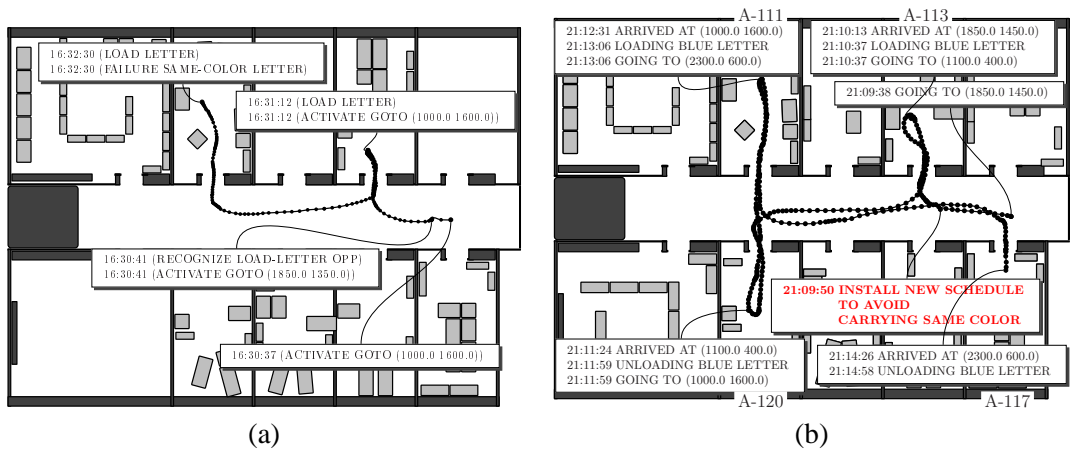

(a)                                                    (b)

Figure 6: The trajectory without prediction-based plan revision (Sub-figure (a)) fails because the courier did not foresee the possible complications with loading the second letter. Sub-figure (b) shows a trajectory where the possible flaw is forestalled by the planning mechanism.

Plan projection is identical to plan execution with two exceptions. First, whenever the plan projector interprets a *wait for* or *whenever* it records the corresponding fluents as active triggering conditions. This way, the plan projection mechanism can automatically generate percepts when continuous control processes or exogenous events make the triggering conditions true. For example, when the navigation plan is waiting for the robot to enter the hallway the plan projector probabilistically guesses when the robot motion causes the respective triggering condition to become true. For this time instant, the plan projector generates a sensor input event with the corresponding sensor reading.

Plan projection also differs from plan interpretation in that whenever the robot interacts with the real world, the projected robot must interact with the symbolic representations of the world. The places where this happens are the low-level plans. Thus instead of executing a low-level plan the projector guesses the results of executing these plans and asserts their effects in the form of propositions to the timeline. There are three kinds of effects that are generated by the interpretation of low-level plans: (1) physical changes, such as the robot changing its position, (2) the low-level plan changing the dynamical state of the robot, such as the direction the robot is heading to, and (3) computational effects, such as changing the values of program variables or signalling the success and failure of control routines. Thus the model of a low-level plan used for plan projection is a probability distribution over the sequence of events that it generates and the delays between the subsequent events.

Thinking procedurally, the plan projector works as follows. It iteratively infers the occurrence of the next event until the given plan is completely interpreted. The next event can either be the next event that the low-level plan generates if the computational state of the controller does not change, or a sensor input event if an active triggering condition is predicted to become true, or an exogenous event if one is predicted to occur. The next predicted event is the earliest of these events.

We will now consider a particular instance of low-level plans: the low-level navigation plans used in the example of the previous section. Navigation is a key action of autonomous mobile robots. While predicting the path that a robot will take it is necessary to predict *where* the robot will be, which is a prerequisite for predicting *what* the robot will be able to perceive. For example, whether the robot will perceive that a door is open depends on the robot taking a path that passes by the door, executing the door angle estimation routine while passing the door and the door being within sensor range. Passing the door is perceived based on the robot's position estimate and the environment map. Consequently, if the robot executes a plan step only if a door is open, then in the end the execution of this plan step depends on the actual path the robot will take. This implies that an action planning process must be capable of predicting the trajectory accurately enough to predict the global course of action correctly.

Navigation actions are representative for a large subset of physical robot actions: they are movements controlled by motors. Physical movements have a number of typical characteristics. First, they are often inaccurate and unreliable. Second, they cause continuous (and sometimes discontinuous) change of the respective part of the robot's state. Third, the interference of concurrent movements can often be described as the superposition of the individual movements.

To discuss the issues raised by the projection of concurrent reactive plans, we sketch a delivery tour plan that specifies how a robot is to deliver mail to the rooms A-113, A-111, and A-120 in Figure 1 (Beetz, 2001). The mail for room A-120 has to be delivered by 10:30 (a strict deadline). Initially, the planner asks the robot to perform the deliveries in the order A-113, A-111, and A-120. As the room A-113 is closed the corresponding delivery cannot be completed. Therefore, the planning system revises the overall plan such that the robot is to accomplish the delivery for A-113 as an opportunity. In other words, the robot will interrupt its current delivery to deliver the mail to A-113 (see Figure 7) if the delivery can be completed.

<u>with policy</u> <u>as long as</u> **in-hallway?**
     <u>whenever</u> **passing-a-door?**
      ESTIMATE-DOOR-ANGLE*()*
<u>with policy</u> <u>seq</u> <u>wait for</u> **open?(A-113)**
      DELIVER-MAIL-TO(DIETER)
   *1.* GO-TO(A-111)
   *2.* GO-TO(A-120) <u>before</u> *10:30*

Figure 7: Delivery tour plan with a concurrent monitoring process triggered by the continuous effects of a navigation plan (passing a door) and an opportunistic step. This concurrent reactive plans serve as an example for discussing the requirements that the causal models must satisfy.

The plan contains constraining sub-plans such as "whenever the robot passes a door it estimates the opening angle of the door using its laser range finders" and opportunities such as "complete the delivery to room A-113 as soon as you learn the office is open". These sub-plans are triggered or completed by the continuous effects of the navigation plans. For example, the event passing a door occurs when the robot traverses a rectangular region in front of the door. We call these events *endogenous events*.

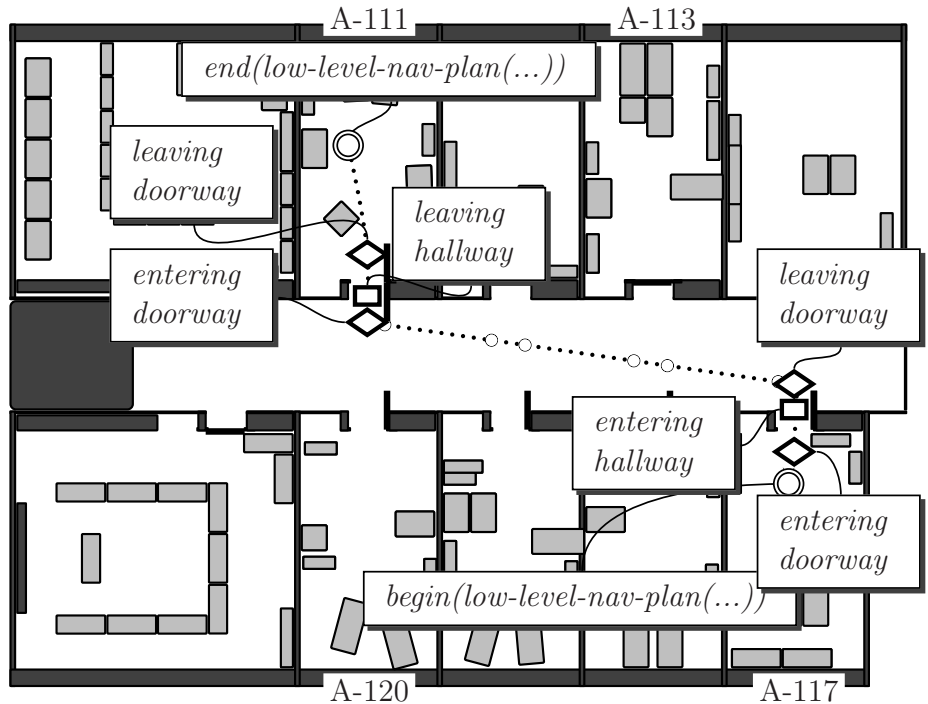

Figure 8: Visualization of a projected execution scenario. The following types of events are depicted by specific symbols: change travel mode event by rhombuses, start/stop passing doorway by small circles, start/stop low-level navigation plan by double circles, and entering doorway/hallway by boxes.

Figure 8 shows a projected execution scenario for a low-level navigation plan embedded in the plan depicted in Figure 7. The behavior generated by low-level navigation plans is modeled as a sequence of events that either cause qualitative behavior changes (e.g. adaptations of the travel mode) or trigger conditions that the plan is reacting to (e.g. entering the hallway or passing a door). The events depicted by rhomboids denote events where the CRP changes the direction and the target velocity of the robot. The squares denote the events entering and leaving offices. The small circles denote the events *starting* and *finishing passing a door*, which are predicted because a concurrent monitoring process estimates the opening angles of doors *while* the robot is passing them.

Such projected execution scenarios have been used for prediction-based debugging of delivery tours of an autonomous robot office courier. Beetz et al. (1999) have shown that a controller employing predictive plan scheduling using the causal models described in this article can perform better than it possibly could without predictive capabilities (see also Section 6.1).

## 2.3 Peculiarities of Projecting Concurrent Reactive Plans

There are several peculiarities in the projection of concurrent reactive plans that we want to point out here.

**Continuous Change.** Concurrent reactive plans activate and deactivate control processes and thereby cause continuous change of states such as the robot's position. Continuous change must be represented explicitly because CRPs employ sensing processes that continually measure relevant states (for example, the robot's position) and promptly react to conditions caused by the continuous effects (for example, entering an office).

**Reactive Control Processes.** Because of the reactive nature of robot plans, the events that have to be predicted for a continuous navigation process depend not only on the process itself but also on the monitoring processes that are simultaneously active and wait for conditions that the continuous effects of the navigation process might cause. Suppose the robot controller is running a monitoring process that stops the robot as soon as it passes an open door. In this case the planner must predict "robot passes door" events for each door the robot passes during a continuous navigation action. These events then trigger a sensing action that estimates the door angle, and if the predicted percept is an "open door detected" then the navigation process is deactivated. Other discrete events that might have to be predicted based on the continuous effects of navigation include entering and leaving a room, having come within one meter of the destination, etc.

**Interference between continuous effects.** For the control processes that set voltages for the robot's motors, the possible modes of interference between control processes are limited. If they generate signals for the same motors the combined effects are determined by the so-called *task arbitration scheme* (Arkin, 1998). The most common task arbitration schemes are (1) behavior blending (where the motor signal is a weighted sum of the current input signals) (Konolige, Myers, Ruspini, & Saffiotti, 1997); (2) prioritized control signals (where the motor signal is the signal of the process with the highest priority) (Brooks, 1986); and (3) exclusion of concurrent control signals through the use of semaphores. In our plans, we exclude multiple control signals to the same motors but they can be easily incorporated in the prediction mechanism. Thus the only remaining type of interference is the superposition of movements such as turning the camera while moving.

**Uncertainty.** There are various kinds of uncertainty and non-determinism in the robot's actions that a causal model should represent. It is often necessary to specify a probability distribution over the average speed and the displacements of points on the paths to enable models to predict the range of spatio-temporal behavior that a navigation plan can generate. Another important issue is to model probability distributions over the occurrence of exogenous events. In most dynamic environments exogenous events such as opening and closing doors might occur at any time.

## 3. Modeling Reactive Control Processes and Continuous Change

Let us now conceptualize the behavior generated by modern robot plans and the interaction between behavior and the interpretation of reactive plans. We base our conceptualization on the vocabulary of *hybrid systems*. Hybrid systems have been developed to design, implement, and verify embedded systems, collections of computer programs that interact with each other and an analog environment (Alur, Henzinger, & Wong-Toi, 1997; Alur, Henzinger, & Ho, 1996).

The advantage of a hybrid system based conceptualization over state-based ones is that hybrid systems are designed to represent concurrent processes with interfering continuous effects. They also allow for discrete changes in process parameterization, which we need to model the activation, deactivation, and reparameterization of control processes through reactive plans. In addition, hybrid system based conceptualizations can model the procedural meaning of *wait for* and *whenever* statements.

As pictured in Figure 9, we consider the robot and its operating environment as two interacting processes: the environment including the robot hardware, which is also called the controlled process, and the concurrent reactive plan, which is the controlling process. The state of the environment is represented by state variables including the variables $x$ and $y$, the robot's real position and $door\text{-}angle_i$ representing the opening angle of door $i$. The robot controller uses fluents to store the robot's measurements of these state variables (*robot-x*, *robot-y*, *door-a120*, etc.). The fluents are continually updated by the self-localization process and a model-based estimator for estimating the opening angles of doors. The control inputs of the plan for the environment process is a vector that includes the *travel mode*, the parameterization of the navigation processes and the current target point to be reached by the robot.

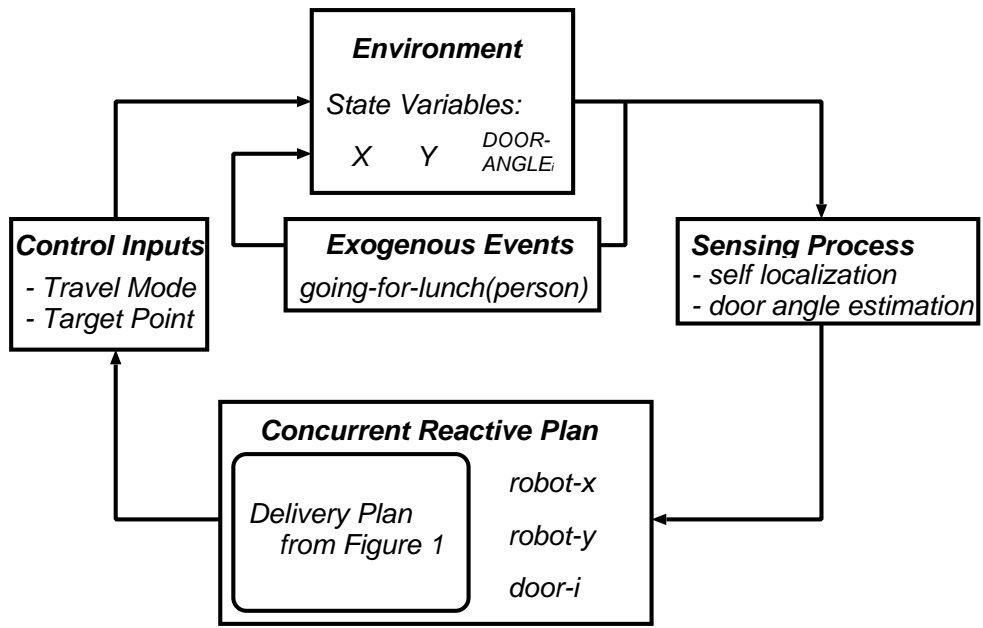

Figure 9: The figure shows our conceptualization of the execution of navigation plans. The relevant state variables are the x and y coordinates of the robot's position and opening angles of the doors. The fluents that estimate these state variables are *robot-x*, *robot-y*, and *door-a120*.

## 3.1 A Hybrid System Model for Reactive Robot Behavior

We will now model the controlled process as a hybrid system (Alur et al., 1997, 1996). Hybrid systems are continuous variable, continuous time systems with a phased operation. Within each phase, called *control mode*, the system evolves continuously according to the dynamical law of that mode, called *continuous flow*. Thus the state of the hybrid system can be thought of as a pair — the control mode and the continuous state. The control mode identifies a flow, and the continuous flow identifies a position in it. Also associated with each control mode are so-called *jump conditions*, specifying the conditions that the discrete state and the continuous state together must satisfy to enable a transition to another control mode. The transitions can cause abrupt changes of the discrete as well as the continuous state. The *jump relation* specifies the valid settings of the system variables that might occur during a jump. Then, until the next transition, the continuous state evolves according to the flow identified by the new control mode.

When considering the interpretation of concurrent reactive plans as a hybrid system the control mode is determined by the set of active control processes and their parameterization. The continuous state is characterized by the system variables $x$, $y$, and $o$ that represent the robot's position and orientation. The continuous flow describes how these state variables change as a result of the active control processes. This change is represented by the component velocities $\dot{x}$, $\dot{y}$, and $\dot{o}$. Thus for each control mode the robot's velocity is constant. Linear flow conditions are sufficient because the robot's paths can be approximated accurately enough using polylines (Beetz & Grosskreutz, 1998). They are also computationally much easier and faster to handle. The jump conditions are the conditions that are monitored by constraining control processes which activate and deactivate other control processes.

Thus the interpretation of a navigation plan according to the hybrid systems model works as follows. The hybrid system starts at some initial state $\langle cm_0, x_0 \rangle$. The state trajectory evolves with the control mode remaining constant and the continuous state $x$ evolving according to the flow condition of $cm$. When the (estimated) continuous state satisfies the transition condition of an edge from mode $cm$ to a mode $cm'$ a jump must be made to mode $cm'$, where the mode might be chosen probabilistically. During the jump the continuous state may get initialized to a new value $x'$. The new state is the pair $\langle cm', x' \rangle$. The continuous state $x'$ evolves according to the flow condition of $cm'$.

The construction of the hybrid system for a given concurrent plan is straightforward. We start at the current plan execution state. For every concurrent active statement of the form ***wait for*** *cond* and ***whenever*** *cond* we add *cond* as a jump condition to the current control mode. In addition we have one additional jump condition for the completion of the plan step.

Figure 10 depicts the interpretation of the first part of the navigation plan shown in Figure 2. The interpretation is represented as a tree where the nodes represent the control modes of the corresponding hybrid system and the node labels the continuous flow. The edges are the control mode transitions labeled with the jump conditions. The robot starts executing the plan in room A-117. The initial control mode of the hybrid system is the root of the state tree depicted in Figure 10. The initial state represents the state of computation where the first control processes of the two parallel branches are active, that is the processes for going to the intermediate target point **1** and maintaining the "office mode" as the robot's travel mode. The flow specifies that while the robot is in the initial control mode the absolute value of the derivative of the robot's position is a function of the robot's navigation mode (office, doorway, or hallway) and the next intermediate target point. The hybrid system makes a transition into one of the subsequent states when either the first target point is reached or when the distance to the doorway becomes less than one meter. The transition condition for the upper edge is that the robot has come sufficiently close to the doorway, for the lower edge that it has reached the first target point. For the lower edge, the hybrid system goes into the state where the robot goes to the target point **2** while still keeping the office mode as its current travel mode. In the other transition the robot changes its travel mode to doorway and keeps approaching the first target point. The only variables that are changed through the control mode transitions are the velocity of the robot and its orientation. Both settings are implied by the flow condition of the respective successor states.

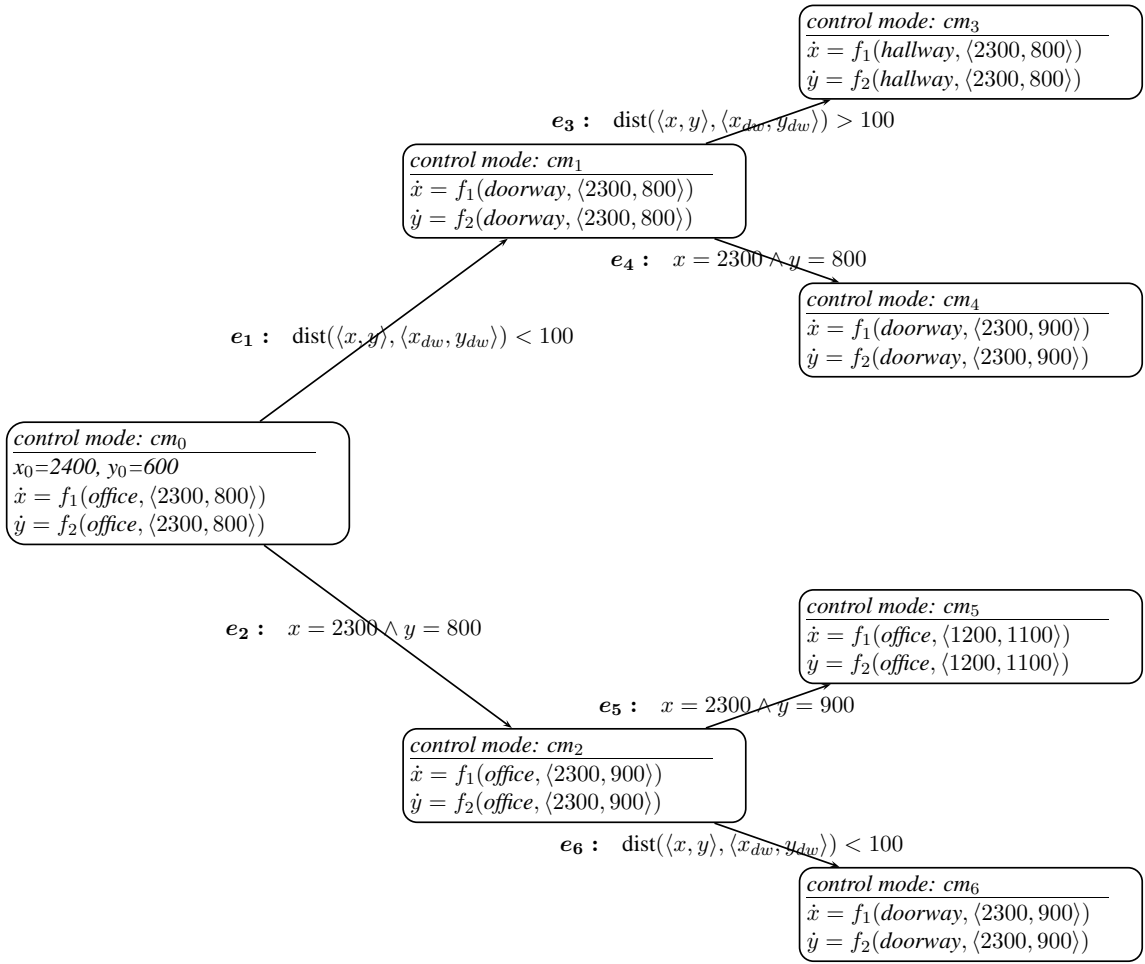

Figure 10: The figure shows the hybrid automaton for the interpretation of the navigation plan in Figure 2. The possible control modes with the continuous flow equations are depicted as nodes and the mode transitions as edges. The edges are labeled with jump conditions: entering doorway ($e_1$, $e_6$), leaving doorway ($e_3$), reaching first waypoint ($e_2$, $e_4$), and reaching second waypoint ($e_5$).

There is one issue that we have not yet addressed in our conceptualization: uncertainty. We model uncertainty with respect to the continuous effects and the achievement of jump conditions using multiple alternative successor modes with varying flows and jump conditions. We associate a probability of occurrence with each mode transition. This way we can, for example, represent rotational inaccuracies of navigation actions that are typical for mobile robots.

## 3.2 Representation of the Hybrid System Model

Let us now formalize our hybrid system conceptualization using a logical notation. To do so, we are going to use the following predicates to describe the evolution of the system states: *jump-Condition(cm,e,c)*, *jumpSuccessor(e,cm',probRange)*, *jumpRelation(cm', $\vec{vals}$, $\vec{flows}$)*, and *probRange(e,max)*. *jumpCondition(cm,e,c)* represents mode *cm* being left along edge *e* when condition *c* becomes true. *jumpSuccessor(e,cm',probRange)* defines the non-deterministic successor states *cm'* and the probability *ProbRange* with which they are entered if the system makes a transition along *e*. *jumpRelation(cm', $\vec{vals}$, $\vec{flows}$)* defines the initial values of the state variables and the flow conditions upon entering state *cm'*. A jump *e* causes the automaton to transit probabilistically into a successor mode.

For each possible successor we define a probability range *probRange*. For reasons that are explained below we represent the probability ranges such that they are non-overlapping, their relative sizes are proportional to the probability they represent (the sum of the ranges is 1) and that their boundaries have the form $\frac{i}{2^n}$, where $i$ and $n$ are integers. The predicate *probRange(e,$2^n$)* defines the sum of the ranges. A possible transition with a probability range $[\frac{i}{2^n}, \frac{j}{2^n}]$ is represented as *jumpSuccessor(e,cm',[i,j])*. The predicate *jumpRelation(cm',$\vec{vals}$, $\vec{flows}$)* means that upon entering control mode *cm'* the system variables and flows are initialized as specified by $\vec{vals}$ and $\vec{flows}$.

Using the predicates introduced above, we can state a probabilistic hybrid automaton (Figure 10) for the interpretation of our navigation plan using the following facts.

> *jumpRelation(cm$_0$,$\langle$2400,600$\rangle$, $\langle f_1$(office, $\langle$2300, 800$\rangle$), $f_2$(office, $\langle$2300, 800$\rangle$)$\rangle$)*
> *jumpCondition(cm$_0$,e$_1$,dist($\langle x,y\rangle$, $\langle x_{dw},y_{dw}\rangle$) < 100)*
> *jumpCondition(cm$_0$,e$_2$,x = 2300 ∧ y = 800)*
>
> *jumpSuccessor(e$_1$,cm$_1$,[1,1])*
> *probRange(e$_1$,1)*
>
> *jumpRelation(cm$_1$,$\langle \rangle$,$\langle f_1$(doorway, $\langle$2300, 800$\rangle$), $f_2$(doorway, $\langle$2300, 800$\rangle$)$\rangle$)*
> *jumpRelation(cm$_2$,$\langle \rangle$,$\langle f_1$(office, $\langle$2300, 900$\rangle$), $f_2$(office, $\langle$2300, 900$\rangle$)$\rangle$)*
> ...

The robot starts at position $\langle 2400, 600\rangle$ in control mode $cm_0$ in which the robot leaves the lower office on the right. In this control mode the robot moves with $\langle f_1$(office, $\langle 2300, 800\rangle$), $f_2$(office, $\langle 2300, 800\rangle$)$\rangle$. The navigation system leaves control mode $cm_0$ when coming close to the door (dist($\langle x,y\rangle$, $\langle x_{dw}, y_{dw}\rangle$) < 100) by performing the transition $e_1$. If the system performs the transition $e_1$ then the control flow changes because the low-level navigation plan switches into the navigation mode *doorway*. In our example, this transition is deterministic.

To account for uncertainty in control we make these transitions probabilistically. Thus we can substitute the control mode $cm_1$ by multiple control modes, say $cm_1'$ and $cm_1''$ where the control flows of the modes are sampled from a probability distribution. We can then state for example that with a probability of 75% ($\frac{12}{16}$) the system transits into control mode $cm_1'$ and with 25% ($\frac{4}{16}$) into the mode $cm_1''$ by defining the effects of the transition $e_1$ as follows:

> *jumpCondition(cm$_0$,e$_1$,dist($\langle x,y\rangle$, $\langle x_{dw},y_{dw}\rangle$) < 100)*
> *jumpSuccessor(e$_1$,cm$_1'$,[1,12])*
> *jumpSuccessor(e$_1$,cm$_1''$,),[13,16])*
> *probRange(e$_1$,16)*

To represent the state of a hybrid automaton we use the predicates *mode(cm)* and *startTime(cm,t)* to represent that the current control mode is *cm* and that *cm* started at time *t*. We use *flow(flow⃗)* and *valuesAt($t_i, \vec{val}_i$)* to assert the flows and values of system variables for given time points. Further, the values of system variables can be inferred for arbitrary time points through interpolation on the basis of the current flow and the last instances of *valuesAt($t_i, \vec{val}_i$)*. This is done using the predicate *stateVarsVals*:

$$stateVarVals(\vec{vals}) \equiv \text{valuesAt}(t_0, \vec{vals}_0) \wedge now(t)$$
$$\wedge flow(\vec{flow}) \wedge \vec{vals} = \vec{vals}_0 + (t - t_0)\vec{flow}$$

where *now(t)* specifies that *t* is the current time. Note, in this conceptualization we represent the discrete state changes explicitly and the states within a mode using the mode's initial state and its flow. A particular state within a mode can be derived on demand using the predicate *stateVarVals*. Interferences between different movements of the robot issued in different control threads are modeled through the mode's flow.

Figure 11 depicts an execution scenario, a possible evolution of the hybrid system representing how the execution of a robot controller might go. An execution scenario is a consistent set of jumps and values from the hybrid model over time. From this we can extract event histories that can be used to simulate plan execution and look for flaws.

An execution scenario consists of a *timeline*, a linear sequence of events and their results. Timelines represent the effects of plan execution in terms of *time instants*, *occasions*, and *co-occurring events*. This implies that several events can occur at the same time instant but only one of them, the primary one, changes the state of the world. *Time instants* are points in time at which the world changes due to an action of the robot or an exogenous event. Each time instant has a *date* which holds the time on the global clock at which the time instant occurred. An *occasion* is a stretch of time over which a world state *P* holds and is specified by a proposition, which describes *P*, and the time interval for which the proposition is true.

We deal with other kinds of uncertainty by representing our model using McDermott's rule language for probabilistic, totally-ordered temporal projection (McDermott, 1994). Using this language we can represent Poisson distributed exogenous events, probability distributions over the current world state, and probabilistic sensor and action models in a way that is consistent with our model presented so far.

### 3.3 Discussion of the Model

Let us now discuss how our hybrid system model addresses the issues raised in Section 2.3. There are two inference tasks concerning the issue of continuous change caused by concurrent reactive plans that are supported by our model. The first one is inferring the state at a particular time instant. For example, if the projection mechanism predicts the occurrence of an exogenous event, such as an object falling out of the robot's gripper, then the projection mechanism has to infer where the robot is at this time instant to assert the position of the object after falling down. This can be done using the initial state and the flow condition of the active control mode. The second important inference task is the prediction of control mode jumps caused by the continuous effects of low-level plans, such as the robot entering the hallway. This can be inferred using the jump conditions of the active control mode in addition to the initial state and the flow condition.

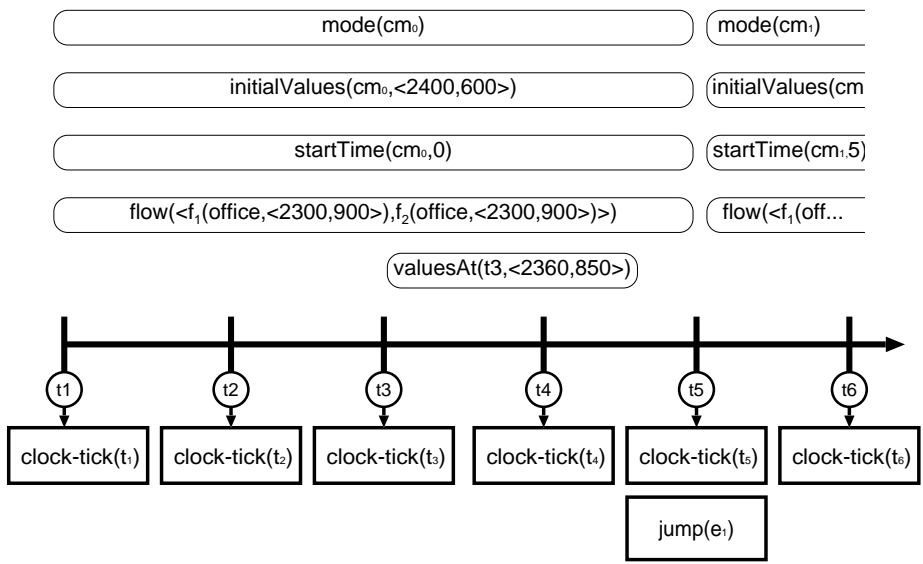

Figure 11:  Part of a timeline that represents a projected execution scenario for a low-level navigation plan.  Time instants are depicted as circles, events as rectangles, and occasions as rectangles with round corners.

The second issue that we have raised in Section 2.3 is the prediction of the robot's reactions to instantaneous events, such as dropping an object.  Typically, a reactive plan for carrying an object contains sub-plans that ask the robot to stop and pick up the object again as soon as the sensed force in the gripper drops.  These kinds of reactions are handled by checking all active jump conditions immediately after an instantaneous event has occurred.

The third issue in projecting concurrent reactive plans is the interference between simultaneous continuous effects.  In our model, interference is modelled by describing the effects of control mode jumps on the flow condition of the subsequent control mode.  The programmer must specify rules describing the physics of the domain for specifying the flow condition of the next mode.  Thus, when a sub-plan for moving the robot's arm is started while the robot is moving, then the rule describing the effects of the corresponding control mode jump asserts the flow condition specifying that the world coordinates of the gripper are determined by the gripper position at the mode jump and the transposition of the two motions in the successor mode.

Our last issue is that of uncertainty.  One aspect of uncertainty that our model supports are inaccuracies in the physical behavior of the robot.  This is modelled by specifying probability distributions over the successor modes when a control mode jump occurs.  Other aspects of uncertainty including probabilistic sensor models, uncertainty of instantaneous physical effects, uncertainty about the state of the world, and Poisson distributed exogenous events are handled by the rule language that we describe in the next section.  In particular, we will give examples of exogenous events and passive sensors in Section 5.3 and a detailed probabilistic model of a complex sensing action in Section 5.4.

Our approach does not explicitly reason about the belief state. We assume that the belief state is computed by probabilistic state estimators (Thrun et al., 2000). Such state estimators not only return the most likely state but also infer additional properties of the belief state such as its ambiguity and the expected accuracy of its global maximum. The plan-based controller interrupts delivery tours as soon as the position estimate is ambiguous or too inaccurate. Details of this mechanism as well as motivations for it can be found in the work of Beetz, Burgard, Fox, and Cremers (1998).

## 4. Probabilistic, Totally-Ordered Temporal Projection

In the last section we have seen how hybrid systems and execution scenarios are represented. In this section, we will see how we can predict execution scenarios from the specification of a hybrid system. For this purpose we use McDermott's rule language for probabilistic, totally-ordered temporal projection (McDermott, 1994). This rule language has the expressiveness needed for our purpose: we can specify the probabilities of event effects depending on the respective situation, Poisson distributed events, and probability distributions over the delays between subsequent plan generated events. Uncertainty about the current state of the world can be specified in the form of probabilistic effect rules of the distinct start event.

This rule language is an excellent basis for formalizing our model introduced in the last section. A set of rules that satisfies certain conditions implies a unique distribution of dated event sequences that satisfies the probabilistic conditions of the individual rules (see Definition 2 in Section 4.1). Thus if we give probabilistic formalizations of the behavior within control modes and mode jumps in McDermott's rule language then we define a unique probability distribution over the state trajectories of the hybrid automaton that satisfies our probabilistic constraints. Moreover, McDermott has developed a provably correct projection algorithm that samples dated event sequences from this unique distribution.

In the remainder of this section we will proceed as follows. We start by presenting McDermott's rule language for probabilistic, totally-ordered temporal projection and summarize the main properties of this language. We will then represent our hybrid system model using this rule language. Based on this representation we can, loosely speaking, show that when applying McDermott's projection algorithm to the representation of the hybrid system, the algorithm returns dated event sequences drawn from the unique distribution implied by our rules with arbitrarily high probability.

Note, to obtain these results we will use a discretized model of time with clock tick events that can be spaced arbitrarily close together resulting in higher accuracy of the projection algorithm. This makes the use of this representation infeasible in practice. Therefore, we eliminate the need for clock tick events in Section 5 by making use of McDermott's *persist* effects.

### 4.1 McDermott's Rule Language for Probabilistic, Totally-ordered Temporal Projection

The different kinds of rules provided by this language are projection rules, effect rules, and exogenous event rules. Projection rules specify the sequence of dated events caused by low-level plans, effect rules specify causal models of sensing processes and actions, and exogenous event rules are used to specify the occurrence of events not under the control of the robot. We will describe these kinds of rules below.

- *Projection rules* can be used to specify the sequence of events caused by the interpretation of a low-level plan. Projection rules have the form

> _project rule_ name(args)
> _if_ cond
> _with a delay of_ $\delta_1$ _occurs_ $ev_1$
> ...
> _with a delay of_ $\delta_n$ _occurs_ $ev_n$

and specify that if low-level plan _name(args)_ is executed and condition _cond_ holds then the low-level plan generates the events $ev_1$, ..., $ev_n$ with relative delays $\delta_1$, ..., $\delta_n$, respectively. Thus, projection rules generate a sequence of dated events.

Uncertain models can be represented by sampling the $\delta_i$ from a probability distribution over durations and by specifying conditions that are satisfied only with a certain probability.

- **_Effect rules_** are used to specify conditional probabilistic effects of events. They have the form

> _e→p rule_ name
> _if_ cond
> _then_ _with probability_ $\theta$ _event_ ev _causes_ effs and specify that whenever event _ev_ occurs and _cond_ holds, then with probability $\theta$ create and clip states as specified in _effs_. The effects of the _e→p rule_ rules have the form $A$, causing the occasion $A$ to hold, _clip_ $A$, causing $A$ to cease to hold, and _persist_ $t$ $A$, causing $A$ to hold for $t$ time units.

- **_Exogenous event rules_** are used to specify the conditional occurrence of exogenous events. The rule

> _p→e rule_ name
> _while_ cond _with an avg spacing of_ $\tau$ _occurs_ ev specifies that over any interval in which _cond_ is true, exogenous events _ev_ are generated Poisson-distributed with an average spacing of $\tau$ time units.

Before proving properties of our model we must first introduce McDermott's semantics of possible worlds. To do so, we define the key notions of the underlying conceptualization in Definition 1. The evolution of the world is described as a sequence of dated instantaneous events where an occurrence $e@t$ specifies that event $e$ occurs at time instant $t$. In addition, we have a function mapping time instants into world states. More precisely these notions are defined as follows (see McDermott, 1994).

**Definition 1** _A **world state** is a function from propositions to $\{T, F, \bot\}$ and is extended to boolean formulas in the usual way. An **occurrence** $e@t$ is a pair $c = (e, t)$, where $e$ is an event and $t$ a time ($t \in \Re_+$). An **occurrence sequence** is a finite sequence of occurrences, ordered by date. Its duration is the date of the last occurrence. A **world of duration L**, where $L \in \Re_+$, is a complete history of duration $L$, that is, it is a pair $(C, H)$, where $C$ is an occurrence sequence with duration $\leq L$, and $H$ is a function from $[0, L]$ to world states. In $H(0)$ all propositions are mapped to F, and if $t1 < t2$ and $H(t1) \neq H(t2)$ then there must be an occurrence $e@t$ with $t1 \leq t \leq t2$._

We use the following abbreviations: $(A \downarrow t)(W)$, "A after t in execution scenario W", to mean that there is some $\delta > 0$ such that $\forall t' : t < t' < t + \delta \Rightarrow H(t')(A) = T$. $(A \uparrow t)(W)$, "A before t in W" is similarly defined, but the upper bound for $t'$ includes t.

After we have described how we represent the change in the world, we state the conditions under which worlds of duration are consistent with a given set of event effects and exogenous event rules. To do so, we have to state the constraints that the rules, the local probabilistic models, impose on state evolution. For this definition we take the plan generated events that are typically specified through project rules, as given.

**Definition 2** *If $T$ is a set of rules as defined above, $Exog$ is an occurrence sequence[2], $P$ is the set of propositions, and $L$ is a real number $\geq duration(Exog)$, then an **L-Model of $T$ and $Exog$** is a pair $(U, M)$, where $U$ is a set of worlds of duration $L$ such that $\forall (C, H) \in U : Exog \subset C$, and $M$ is a probability measure on $U$ that obeys the following restrictions: $(A{\downarrow}t)$, $(A{\uparrow}t)$ and $e@t$ are considered as random variables. $\overline{A}$ is the "annihilation" of a conjunction A, that is the conjunction of the negations of the conjuncts of A.*

1. **Initial blank state:** $\forall A \in P : M(A \uparrow 0) = 0$.

2. **Event-effect rules:** *If $T$ contains a rule instance*

   <u>e→p rule</u>  *name* <u>*if*</u>  *A* <u>*then*</u>  <u>with probability</u> *r* <u>event</u> *e* <u>causes</u> *B,*

   *then for every date t, require that, for all nonempty conjunctions C of literals from B:*
   $M(C{\downarrow}t | e@t \land A{\uparrow}t \land \overline{B}{\uparrow}t) = r.$

3. **Event-effect rules when the events don't occur:** *Suppose B is an atomic formula, and let $R = \{R_i\}$ be the set of all instances of <u>e→p</u> rules whose consequents contain B or $\neg B$. If <u>e→p rule</u> $R_i =$ <u>*if*</u> $A_i$ <u>*then*</u> <u>with probability</u> $p_i$ <u>event</u> $E_i$ <u>causes</u> $C_i$, then let $D_i = A_i \land \overline{C_i}$, <u>Then</u> $M(B{\downarrow}t | B{\uparrow}t \land N) = \overline{1}$ and $M(B{\downarrow}t | \neg B{\uparrow}t \land N) = 0$ where $N = (\neg E_1@t \lor \neg D_1) \land (\neg E_2@t \lor \neg D_2) \land \dots$.*

4. **Event-occurrence rules:** *For every time point t such that no occurrence with date t is in $Exog$ and every event E, such that there is exactly one instance*

   <u>p→e rule</u>  *name* <u>while</u>  *A* <u>with an avg spacing of</u>  *d* <u>occurs</u>  *E*

   *with $M(a{\uparrow}t) > 0$ require*

   $$lim_{dt \to 0} \frac{M(\text{occ. of class E between t and t+dt} | A{\uparrow}t)}{dt} = 1/d$$

   $$lim_{dt \to 0} \frac{M(\text{occ. of class E between t and t+dt} | \neg A{\uparrow}t)}{dt} = 0$$

---

2. *Exog* is an occurrence sequence, which represents the events generated by the interpretation of the robot's plan and modeled using projection rules.

*if there exists no so such rule, require*

$$lim_{dt \to 0} \frac{M(occ.\ of\ class\ E\ between\ t\ and\ t+dt)}{dt} = 0$$

5. **Conditional independence:** *If one of the previous clauses defines a conditional probability $M(\alpha|\beta)$, which mentions times t, then $\alpha$ is conditionally independent, given $\beta$, of all other random variables mentioning times on or before t. That is, for arbitrary $\gamma$ mentioning times on or before t, $M(\alpha|\beta) = M(\alpha|\beta \wedge \gamma)$.*

McDermott (1994) shows that this definition yields a unique probability distribution $M$. He also gives a proof that his projection algorithm draws random execution scenarios sampled from this unique probability distribution implied by the given probabilistic models.

### 4.2 Probabilistic Temporal Rules for PHAMs

In order to predict the evolution of a hybrid system, we specify rules in McDermott's rule language that, given the state of the system and a time *t*, predict the successor state at *t*. To predict the successor state, we must distinguish three cases: first, a control mode transition occurs; second, an exogenous event occurs; and third, the system changes according to the flow of the current mode.

We will start with the rules for predicting control mode jumps. To ensure that mode transitions are generated as specified by the probability distributions over the successor modes, we will use the predicate *randomlySampledSuccessorMode(e,cm)* and realize a random number generator using McDermott's rule language.

*randomlySampledSuccessorMode(e,cm)* $\equiv$
    $probRange(e, max) \wedge randomNumber(n, max)$
    $\wedge jumpSuccessor(e, cm, range) \wedge n \in range$

In order to sample values from probability distributions we have to axiomatize a random number generator that asserts instances of the predicate *randomNumber(n,max)* used above (see Beetz & Grosskreutz, 2000). We do this by formalizing a *randomize* event. McDermott (1994) discusses the usefulness of, and the difficulties in, realizing nondeterministic exclusive outcomes. Therefore in his implementation he escapes to Lisp and uses a function that returns a random element.

**Lemma 1** *At any time point randomNumber has exactly one extension randomNumber(r,max) where r is an unbiased random between $0$ and $max$.*

<u>Proof:</u> Let *max\** be the largest *probRange* extension and *randomBit(i,value)* the i-th random bit. The start event that causes the initial state timeline causes *randomBit(i,0)* $\forall 0 \le i \le \log max^*$. Thereafter, a $randomize$ event is used to sample their value:

<u>**e→p rule**</u> RANDOMIZE
<u>*if*</u> *randomBit(i,val)* $\wedge$ *negation(val,neg)*
<u>*then*</u> <u>*with probability*</u> *0.5*
    <u>*event*</u> *randomize*
    <u>*causes*</u> *randomBit(i,neg)* $\wedge$ <u>*clip*</u> *randomBit(i,val)*

Rule MODE-JUMP causes a control mode transition as soon as the jump condition *cond* becomes true. The rule says that in any interval in which *cm* is the current control mode and in which the jump condition *cond* for leaving *cm* following edge *edge* a jump along *edge* will occur with an average delay of $\tau$ time units.

*p→e rule* MODE-JUMP

*while* *mode(cm)* ∧ *jumpCondition(cm,cond,edge)*
$\qquad$ ∧ *stateVarsVal($\vec{vals}$)* ∧ *satisfies(vals,cond)*

*with an average spacing of* $\tau$ *time units*

*occurs* *jump(edge)*

Rule JUMP-EFFECTS specifies the effects of an jump event on the control mode, system variables, and the flow. If *cm* is a control mode randomly sampled from the probability distribution over successor nodes for jumps along *edge* then the jump along *edge* has the following effects. The values of the state variables and the flow condition of the previous control mode $cm_{old}$ are retracted and the ones for the new control mode *cm* are asserted.

*e→p rule* JUMP-EFFECTS

*if* *randomlySampledSuccessorMode(edge,cm)*
$\qquad$ ∧ *initialValues(cm,$\vec{val}$)* ∧ *flowCond(cm,$\vec{flow}$)* ∧ *now(t)*
$\qquad$ ∧ *mode($cm_{old}$)* ∧ *flow($flow_{old}$)* ∧ *valuesAt($t_{old}$,$val_{old}$)*

*then* *with probability* 1.0

$\qquad$ *event* *jump(edge)*

$\qquad$ *causes* *mode(cm)* ∧ *flow($\vec{flow}$)* ∧ *valuesAt(transTime,$\vec{val}$)*
$\qquad\qquad$ ∧ *clip* *mode($cm_{old}$)* ∧ *clip* *flow($flow_{old}$)* ∧ *clip* *valuesAt($t_{old}$,$val_{old}$)*

Time is advanced using *clock-tick* events. With every CLOCK-TICK(*?t*) event the *now* predicate is updated by clipping the previous time and asserting the new one. Note, the time differs at most $dt_{clock}$ time units from the actual time.

*e→p rule* CLOCK-RULE

*if* *now($t_o$)*

*then* *with probability* 1.0

$\qquad$ *event* *clock-tick(t)*

$\qquad$ *causes* *now(t)* ∧ *clip* *now($t_o$)*

Exogenous events are modeled using rules of the following structure. When the navigation process is in the control mode *cm* and the values *$\vec{vals}$* of the state variables satisfy the condition for the occurrence of the exogenous event *ev*, then the event *ev* occurs with average spacing of $\tau$ time units.

*p→e rule* CAUSE-EXO-EVENT

*while* *mode(cm)* ∧ *exoEventCond(cm,cond,ev)*
$\qquad$ ∧ *stateVarsVal($\vec{vals}$)* ∧ *satisfies(vals,cond)*

*with an average spacing of* $\tau$ *time units*

*occurs* *exoEvent(ev)*

The effects of exogenous event rules are specified by rules of the following form. The exogenous event *exoEvent(ev)* with effect specification *exoEffect(ev,$\vec{val}$))* causes the values of the state variables to change from $\vec{val_o}$ to $\vec{val}$.

*e→p rule*  EXO-EVENT-EFFECT
*if*  *exoEffect(ev,$\vec{val}$))* $\land$ *valuesAt(t$_o$,val$_o$)* $\land$ *now(t)*
*then*  *with probability* 1.0
    *event*  *exoEvent(ev)*
    *causes*  *valuesAt(t,$\vec{val}$)* $\land$ *clip*  *valuesAt(t$_o$,$\vec{val_o}$)*

### 4.3 Properties of PHAMs

We have seen in the last section that a PHAM consists of the rules above and a set of facts that constitute the hybrid automata representation of a given CRP. In this section we investigate whether PHAMs make the "right" predictions.

There are essentially three properties of predicted execution scenarios that we want to ensure. First, predicted control mode sequences are consistent with the specified hybrid system. Second, mode jumps are predicted according to the specified probability distribution over successor modes. Third, between two successive events, the behavior is predicted according to the flow of the respective control mode.

As McDermott's formalism does not allow for modeling instantaneous state transitions we can only show that control mode sequences in execution scenarios are probably approximately accurate. In our view, this is a low price for the expressiveness we gain through the availability of Poisson distributed exogenous events.

The subsequent lemma 2 states that control mode jumps can be predicted with arbitrary accuracy and arbitrarily high probability by decreasing the time between successive clock ticks.

**Lemma 2** *For each probability $\epsilon$ and delay $\delta$, there exists a $\tau$ (average delay of the occurrence of an event after the triggering condition has become true) and a $dt_{clock}$ (time between two subsequent clock ticks) such that whenever a jump condition becomes satisfied, then with probability $\geq 1 - \epsilon$ a jump event will occur within $\delta$ time units.*

**Proof:** Let $t$ be the time where the jump condition is fulfilled. If $\tau \leq \delta/(2\log(1/\epsilon))$ and $dt_{clock} \leq \delta/2$ then at most $\delta/2$ time units after $t$ the antecedent of rule MODE-JUMP is fulfilled. The probability that no event of class $jump(cm')$ occurs between $t+\delta/2$ and $t+\delta$ is $\leq e^{-\delta/(2\tau)} = e^{-log(1/\epsilon)} = \epsilon$, so with probability $\geq 1 - \epsilon$ such an event will occur at most $\delta$ time units after $t$.

$\square$

This implies that there is always a non-zero chance that control mode sequences are predicted incorrectly. It happens only when two jump conditions become true and the jump triggered by the later condition occurred before the other one. However, the probability of such incorrect predictions can be made arbitrarily small by the choice of $\tau$ and $dt_{clock}$.

The basic framework of hybrid systems does not take the possibility of exogenous events into account and thereby allows for proving strong system properties such as the reachability of goal states from arbitrary initial conditions or safety conditions for the system behavior (Alur et al., 1997, 1996). For the prediction of robot behavior in dynamic environments these assumptions, however, are unrealistic. Therefore, we only have a weaker property, namely the correspondence between the predicted behavior and the flows specified by the hybrid system between immediate subsequent events.

**Lemma 3** *Let $W$ be an execution scenario, $e_1@t_1$ and $e_2@t_2$ be two immediate subsequent events of type jump or exoEvent, and cm be the control mode after $t_1$ in $W$. Then, for every occurrence $e@t$ with $t_1 < t \leq t_2$ $W(t)(stateVarVals(\vec{vals}))$ is unique. Further, $\vec{vals} = \vec{vals}_1 + (t - t_1) *$ flow(cm), where $\vec{vals}_1$ are the values of the state variables at $t_1$.*

**Proof:** *There are only two classes of rules that affect the value of valuesAt and flow: rule* JUMP-EFFECTS, *and rule* EXO-EVENT-EFFECT. *These rules always clip and set exactly one extension of the predicates, thus together with the fact that the initial event asserts exactly one such predicate, the determined value is unique.*

*During the interval between $t_1$ and $t_2$ the extension of stateVarVals evolves according to the flow condition of mode $cm$ due to the fact that flow is not changed by rule* EXO-EVENT-EFFECT. *Thus it remains as initially set by rule* JUMP-EFFECTS, *which asserts exactly the flow corresponding to cm. The proposition then follows from the assumption of a correct axiomatization of addition and scalar-vector multiplication.*

$\square$

Another important property of our representation is that jumps are predicted according to the probability distributions specified for the hybrid automaton.

**Lemma 4** *Whenever a jump along an edge $e$ occurs, the successor state is chosen according to the probability distribution implied by probRange and jumpSuccessor.*

**Proof:** *This follows from the properties of the randomize event and Rule Jump-Effects.*

$\square$

Using the lemmata we can state and show the central properties of PHAMs: (1) the predicted control mode transitions correspond to those specified by the hybrid automaton; and (2) the same holds for the continuous predicted behavior between exogenous events; (3) Exogenous events are generated according to their probabilities over a continuous domain (this is shown in McDermott, 1994).

**Theorem 1** *Every sequence of $mode(cm)$ occasions follows a branch $(cm_i), ..., (cm_j)$ of the hybrid automaton.*

**Proof:** *Each occasion mode(cm) must be asserted by rule* JUMP-EFFECTS. *Therefore there must have been a jump(e) event. Consequently, there must have been a jumpCondition from the previous control mode to cm.*

$\square$

Because jump events are modeled as Poisson distributed events there is always the chance of predicting control mode sequences that are not valid with respect to the original hybrid system. So next we will bound the probability of predicting such mode sequences by choosing the parameterization of the jump event and clock tick event rules appropriately.

**Theorem 2** *For every probability $\epsilon$ there exists an average delay of a mode jump event $\tau$ and a delay $dt_{clock}$ with which the satisfaction of jump conditions is realized such that with probability $\geq 1 - \epsilon$ the $\vec{vals}$ of stateVarVals occasions between two immediate subsequent exogenous events follow a state trajectory of the hybrid automaton.*

**<u>Proof:</u>** The proof is based on the property that jumps occur in their correct order with an arbitrarily high probability. In particular, we can choose $\delta$ as a function of the minimal delay between jump conditions becoming true. Then, the jumps to successor modes occur with arbitrarily high probability (Lemma 2). Finally, according to Lemma 3 the trajectory of *stateVarVals* between transitions is accurate.

<div align="right">□</div>

## 5. The Implementation of PHAMs

We have now shown that PHAMs define probability distributions over possible execution scenarios with respect to a given belief state. The problem of using PHAMs is obvious. Nontrivial CRPs for controlling robots reliably require hundreds of lines of code. There are typically several control processes active, many more are dormant, waiting for conditions that trigger their execution. The hybrid automata for such CRPs are huge, the branching factors for mode transitions are immense. Let alone the distribution of execution scenarios that they might generate. The accurate computation of this probability distribution is prohibitively expensive in terms of computational resources.

There is a second source of inefficiency in the realization of PHAMs. In PHAMs we have used clock tick rules, Poisson distributed events, that generate clock ticks with an average spacing of $\tau$ time units. We have done so, in order to formalize the operation of CRPs in a single concise framework. The problem with this approach is that in order to predict control mode jumps accurately we must choose $\tau$ to be very small. This, however, increases the number of clock tick events drastically and makes the approach infeasible for all but the most simple scenarios.

In order to draw sample execution scenarios from the distribution implied by the causal model and the initial state description we use an extension of the XFRM projector (McDermott, 1992b) that employs the RPL interpreter (McDermott, 1991) together with McDermott's algorithm for probabilistic temporal projection (McDermott, 1994). The projector takes as its input a CRP, rules for generating exogenous events, a set of probabilistic rules describing the effects of events and actions, and a (probabilistic) initial state description. To predict the effects of low-level plans the projector samples effects from the probabilistic causal models of the low-level plans and asserts them as propositions to the timeline. Similarly, when the plan activates a sensor, the projector makes use of a model of the sensor and the state of the world as described by the timeline to predict the sensor reading.

In this section we investigate how we can make effective and informative predictions on the basis of PHAMs that can be performed at a speed sufficient for prediction-based online plan revision. To achieve effectiveness we use two means. First, we realize weaker inference mechanisms that are based on sampling execution scenarios from the distribution implied by the causal models and the initial state description. Second, we replace the clock tick event mechanism with a different mechanism that infers the occurrence of control mode jumps and uses the *persist* effect to generate the respective delay. We will detail these two mechanisms in the remainder of this section.

### 5.1 Projection with Adaptive Causal Models

Let us first turn to the issue of eliminating the inefficiencies caused by the clock tick mechanism. We will do so by replacing clock tick rules with a mechanism for tailoring causal models on the fly and using the *persist* effects of the probabilistic rule language.

For efficiency reasons the process of projecting a continuous process *p* is divided into two phases. The first phase estimates a schedule for endogenous events caused by *p* while considering possible effects of *p* on other processes but not the effects of the other processes on *p*. This schedule is transformed into a context-specific causal model tailored for the plan which is to be projected. The second phase projects the plan *p* using the model of endogenous events constructed in the first phase. This phase takes into account the interferences with concurrent events and revises the causal model if situations arise in which the assumptions of the precomputed schedule are violated.

The projection module uses a model of the dynamic system that specifies for each continuous control process the state variables it changes and for each state variable the fluents that measure that state variable. For example, consider the low-level navigation plans that steadily change the robot's position (that is the variables *x* and *y*). The estimated position of the robot is stored in the fluents *robot-x* and *robot-y*:

*changes(low-level-navigation-plan, x)*
*changes(low-level-navigation-plan, y)*
*measures(robot-x, x)*
*measures(robot-y, y)*

**Extracting relevant conditions.** When the projector starts projecting a low-level navigation plan it computes the set of pending conditions that depend on *robot-x* and *robot-y*, which are the fluents that measure the state variables of the dynamic system and are changed by the low-level navigation plan. These conditions are implemented as fluent networks.

Fluent networks are digital circuits where the components of the circuit are fluents. Figure 12 shows a fluent network where the output fluent is true, if and only if the robot is in room A-120. The inputs of the circuit are the fluents *robot-x* and *robot-y* and the circuit is updated whenever *robot-x* and *robot-y* change.

Our reactive plans are set up such that the fluent networks that compute conditions for which the plan is waiting can be determined automatically using (PROLOG-like) relational queries:

*setof*  *?fl-net ( fluent(?fl) ∧ status(?fl,pending)*
          *∧ changes(low-level-nav-plan, ?state-var)*
          *∧ measures(?state-var-fl, ?state-var)*
          *∧ depends-on(?fl, ?state-var-fl)*
          *∧ fluent-network(?fl, ?fl-net) )*
      *?pending-fl-nets*

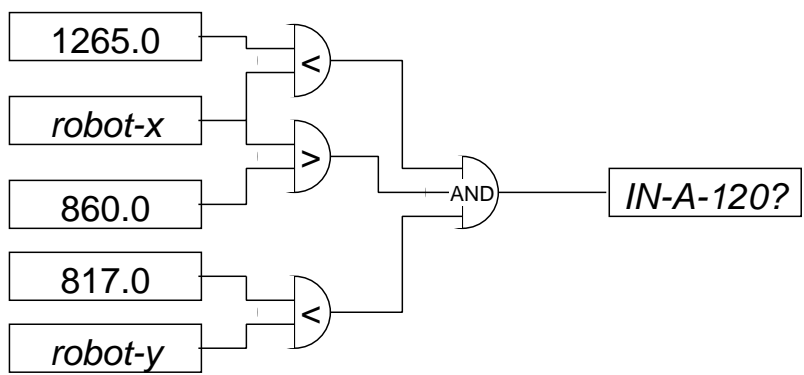

Figure 12: Fluent network for being in room A-120. The robot believes it is in room A-120 if its estimated x-coordinate is between 860 and 1265 and the y-coordinate is smaller than 817, where the hallway begins.

This query determines *?pending-fl-nets*, the set of fluent networks *?fl-net* such that *?fl-net* is a network with output fluent *?fl*. *?fl* causes a plan thread to pend and depends on a fluent measuring a state variable *?state-var* changed by the low-level navigation plan. The extraction of these conditions is done automatically. The automatic extraction requires the conditions be in a particular form and the effects of low-level plans on state variables and the sensing of state variables to be represented explicitly.

To predict when the fluent IN-A-120? will become true or false, we have to compute the region in the state space that corresponds to the fluent and compute the intersections of the robot's state trajectories with this region.

**Endogenous event schedules.** For each class of continuous processes we have to provide an *endogenous event scheduler* that takes the initial conditions and the parameterization of the process, and the fluent networks that might be triggered and computes the endogenous event schedule. The endogenous event scheduler for the low-level navigation plans is described in the next section. Given the kind of process (e.g., low-level navigation plan), the process parameters (e.g., the destination of the robot), and the pending fluent networks, the scheduler returns a sequence of composite endogenous events. Composite events are represented as triples of the form *($\Delta t$, $\langle sv_1, ..., sv_n \rangle$, $\{ev_1, ..., ev_m\}$)*. $\Delta t$ is the delay between the $i$th and the $i+1$st event in the schedule, $\langle sv_1, ..., sv_n \rangle$ the values of the state variables, and $\{ev_1, ..., ev_m\}$ the atomic events that are to take place.

If a state for which the plan is waiting, becomes true at a time instance $t$, then at $t$ a *passive-sensor-update* event is triggered. *passive-sensor-update* is an event model that takes a set of fluents as its parameters, retrieves the values of the state variables measured by these fluents, applies the sensor model to these values, and then sets the fluents accordingly.

**A causal model of low-level navigation plans.** Projecting the initiation of the execution of a navigation plan causes two events: the start event and a hypothetical completion event after an infinite number of time units. This is shown in the following projection rule.

*project rule* LOW-LEVEL-NAVIGATION-PLAN
*if* *true*
*with a delay of* 0
*occurs* *begin(low-level-nav-plan( ?dest-descr, ?id, ?fluent)*
*with a delay of* ∞
*occurs* *end(low-level-nav-plan( ?dest-descr, ?id, ?fluent)*

The effect rule of the start event of the low-level navigation plan computes the endogenous event schedule and asserts the next endogenous navigation event into the timeline.

*e→p rule* ENDOGENOUS-EVENTS
*if* *endogenous-event-schedule(low-level-nav-plan( ?dest-descr, ?schedule))*
*then* *with probability* *1.0*
    *event* *begin(low-level-nav-plan( ?dest-descr, ?id, ?fluent))*
    *causes* *predicted-events( ?id, ?schedule)*
        ∧ *running(robot-goto( ?descr, ?id))*
        ∧ *next-nav-event( ?id))*

The occasion *next-nav-event( ?id)* triggers the next endogenous event *begin(follow-path( ?here ⟨ ?x, ?y⟩) ?dt ?id))*. The remaining two conditions determine the parameters of the *follow-path* event: the next scheduled event and the robot's position.

*p→e rule* CAUSE-EXO-EVENT
*while* *next-nav-event( ?id)*
    ∧ *predicted-events( ?id, (( ?dt ⟨ ?x, ?y⟩ ?evs) ! ?remaining-evs)*
    ∧ *robot-loc( ?here)*
*with an average spacing of* *0.0001*
*occurs* *begin(follow-path( ?here, ⟨ ?x, ?y⟩, ?dt, ?id))*

The effect rule of the *begin(follow-path (...))* event specifies among other things that the next endogenous event will occur after *?dt* time units (*persist* *?dt sleeping( ?id))*.

*e→p rule* FOLLOW-PATH
*if* *robot-loc( ?coords)*
*then* *with probability* *1.0*
    *event* *begin(follow-path( ?from, ?to, ?dt, ?id))*
    *causes* *running(follow-path( ?from, ?to, ?dt, ?id))*
        ∧ *clip* *robot-loc( ?coords)*
        ∧ *clip* *next-nav-event( ?id)*
        ∧ *persist* *?dt sleeping( ?id)*

If a running *follow path* event has finished sleeping the *end (follow-path (...))* event occurs.

*p→e rule* TERMINATE-FOLLOW-PATH
*while* *not* *sleeping( ?id)*
    ∧ *running(follow-path( ?from, ?to, ?time, ?id))*
*with an average spacing of* *0.0001*
*occurs* *end(follow-path( ?from, ?to, ?time, ?id))*

Our model of low-level navigation plan presented so far suffices as long as nothing important happens while carrying out the plan. However, suppose that an exogenous event that causes an object to slip out of the robot's hand is projected at time instant *t* while the robot is in motion. To predict the new location of the object the projector predicts the location *l* of the robot at the time *t* using the control flow and asserts it in the timeline.

Qualitative changes in the behavior of the robot caused by adaptations of the travel mode are described through *e→p* -rules. The following *e→p* -rule describes the effects of the event *nav-event(set-travel-mode(?n))*, which represents the low-level navigation plan resetting the travel mode:

<u>*e→p rule*</u> SET-DOORWAY-MODE
<u>*if*</u> *travel-mode(?m)*
<u>*then*</u> <u>*with probability*</u> *1.0*
     <u>*event*</u> *nav-event(set-travel-mode(doorway))*
     <u>*causes*</u> <u>*clip*</u> *travel-mode(?m)*
          ∧ <u>*clip*</u> *obstacle-avoidance-with(sonar)*
          ∧ *travel-mode(doorway)*

The rule specifies that if at a time instant at which an event *nav-event(set-travel-mode(?n))* occurs the state *travel-mode(?m)* holds for some *?m*, then the states *travel-mode(?m)* and *obstacle-avoidance-with(sonar)* will (with a probability of *1.0*) not persist after the event has occurred, i.e., they are clipped by the event. The event causes the state *travel-mode(doorway)* to hold until it is adapted next time.

The rules listed above are hand-coded and plan-specific. An investigation of whether the plans can be coded such that the rule specification can be automated is on our agenda for future research.

## 5.2 Endogenous Event Scheduler

We have just shown how events are projected from a given endogenous event schedule, but we have not shown how the schedule is constructed. Thus, this section describes the endogenous event scheduler for low-level navigation plans. The scheduler predicts the effects of the low-level navigation plan on the state variables *x* and *y*. The endogenous event scheduler assumes the robot is following a straight path between locations *1* to *5*. As we have pointed out earlier, there are two kinds of events that need to be predicted: the ones causing *qualitative physical change* and the ones causing *trigger conditions* that the plan is waiting for.

The qualitative events caused by the low-level navigation plan pictured in Figure 13 are the ones that occur when the robot arrives at the locations *1*, *2*, *3*, *4*, and *5* in which the robot either changes its travel mode or arrives at its destination. For each of these time instants the occurrence of a *set-travel-mode*-event is predicted.

The scheduler for triggering events works in two phases: (1) it transforms the fluent network into a condition that it is able to predict and (2) it applies an algorithm for computing when these events occur. The conditions that are caused by the low-level navigation plan can be represented as regions in the environment such that the condition is true if and only if the robot is within this region. The elementary conditions are numeric constraints on the robot's position or the distance of the robot to a given target point. The scheduler assumes that *robot-x* and *robot-y* are the only fluents in these networks that change their value during the execution of the plan. More complex networks can be constructed as conjunctions and disjunctions of the elementary conditions.

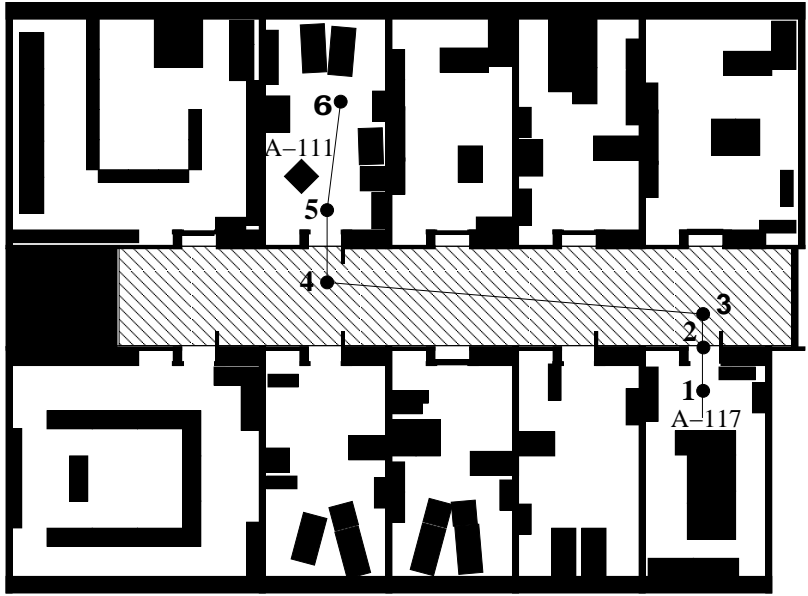

Figure 13: Initially predicted endogenous events.

In the next step the endogenous event scheduler overlays the straight line path through the intermediate goal points of the topological navigation path (see Figure 7) with the regions computed in the previous step. It then computes a schedule for the endogenous events by following the navigation path and collecting the intersections with the regions (see Figure 13). The result of the scheduling step is a sequence of triples of the form $(\Delta t_i, \langle x_i, y_i \rangle, \{ev_1, ..., ev_n\})$.

**Rescheduling endogenous events.** One problem that our temporal projector has to deal with is that a _wait for_ step might be executed while a low-level navigation plan is projected. For example, when the robot enters the hallway, the policy that looks for the opening angles of doors when passing them is triggered. Therefore, the causal model that was computed by the endogenous event scheduler is no longer sufficient. It fails to predict the "passing a door" events.

These problems are handled by modifying the endogenous event schedule: whenever the robot starts waiting for a condition that is a function of the robot's position, it interrupts the projection of the low-level navigation plan, adapts the causal model of the low-level navigation plan, and continues with the projection. In the case of entering the hallway, a new endogenous event schedule that contains endogenous events for passing doorways is computed. This updated schedule of endogenous events is pictured in Figure 14.

### 5.3 Projecting Exogenous Events, Passive Sensors and Obstacle Avoidance

One type of exogenous event is an event for which we have additional information about its time of occurrence, such as the event that Dieter will be back from lunch around _12:25_. These kinds of events are represented by a _p→e_ rule together with an _e→p_ rule. The _e→p_ rule specifies that the

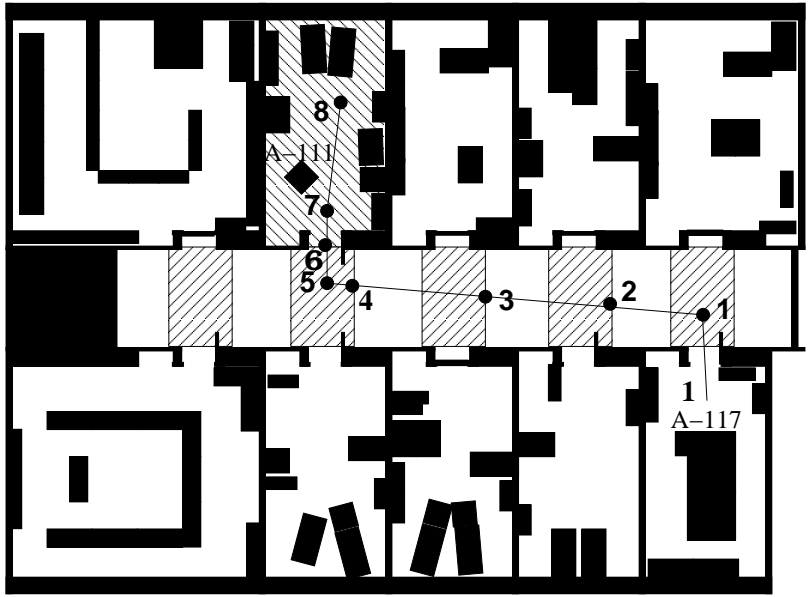

Figure 14: Modified endogenous event schedule.

*start* event causes the state *before-dieters-door-opens()* to hold and persist for *?time* time units. The event *dieters-door-opens()* is triggered as soon as *before-the-door-opens()* no longer holds.

*e→p rule* BACK-FROM-LUNCH
*if* *about(?time, 12:25)* ∧ *difference(?time, *now*, ?wait-for))*
*then* *with probability* *1.0*
    *event* *start*
    *causes* *persist* *?wait-for before-the-door-opens*

*p→e rule* DOOR-OPENS
*while* *thnot* *before-the-door-opens*
*with an average spacing of* *0.0001*
*occurs* *dieters-door-is-opened*

In order to predict the occurrence of exogenous events, the plan projector does the following. It first computes the time when the robot will cause the next event $e_{next}$. Let us assume that this event occurs $t$ time units after the last event $e_{last}$ and that $c$ is the strongest condition that holds from $e_{last}$ until $e_{next}$.[3] The following algorithm predicts the occurrence of the next exogenous event accurately. First, for every *p→e* rule $r_i$ whose enabling condition is satisfied by $c$ randomly decide whether $e_i$ will occur between $e_{last}$ and $e_{next}$ based on the average temporal spacing of $e_i$ events in situations where $c$ holds. If $e_i$ is predicted to occur, select its occurrence time by randomly selecting a time instant in the time interval between $e_{last}$ and $e_{next}$ (the exogenous events are Poisson distributed). Select the exogenous event that is predicted to occur the earliest, assert it to the timeline, and continue the projection after the occurrence of this event.

---

3. The cases where enabling conditions of exogenous events are caused by the continuous effects between $e_{last}$ and $e_{next}$ are handled analogously to the achievement of triggering conditions.

The last two components we need to describe are passive sensors, which are steadily updated and do not change the configuration of the robot and the behavior of the collision avoidance routines.

Readings of passive sensors only need to be projected if the measured state variables change significantly or if the state variables traverse values that satisfy conditions for which the robot is waiting. For each of these situations there is an *update-passive-sensors* event.

Collision avoidance is not modeled except in situations in which the robot is told about objects that are moved around. In this case the endogenous event scheduler adds a region corresponding to the object. If the region blocks the way to the destination — that is the robot cannot move around the region — then a *possible-bump-event* is generated. The effect rule for a possible bump event specifies that, if the robot has activated sensors that can detect the object, the low-level navigation plan fails with a failure description "path blocked." Otherwise a bump event is generated. For example, since sonar sensors are the only sensors placed at table height, the collision avoidance module can avoid a collision with a table only if the sonar sensors are active. Thus, to predict a bump, the projector has to determine how long the sonar sensors have been switched off before the possible bump event occurs.

## 5.4 Models of Complex Sensing Actions

To understand how other types of uncertainty can be modeled and how the causal models interact with the plan interpretation let us look at a more complex sensing action realized through a low-level plan *look-for*. The sub-plan is called with a visual description (*?pl*) of objects it is supposed to look for.

Typically, in order to model a low-level plan we need a set of projection rules that probabilistically describe the possible event sequences and outcomes when activating a behavior module. In each situation exactly one projection rule is applied (although the decision about which one might be probabilistic).

One of these projection rules for *look-for* is listed below. The model consists of three parts. The first part (line 1 to 7) specifies the condition under which this rule predicts the behavior of the *look-for* correctly. The second part (lines 8 to 11) lists the events that *look-for* will cause if this rule is applicable. Finally, the last line specifies how the low-level plan signals the completion of its interpretation. In our case, the low-level plan succeeds and returns a list of object descriptions (*?desigs*) as its value.

The condition of the projection rule determines where the robot is (*1*), probabilistically decides whether the *look-for* is "normal" based on the camera used and the specifics of the location (*2*), and infers what objects are located there (*3*). This inference is performed based on the robot's probabilistic belief about the state of the world and the predicted exogenous events. The condition then uses the sensor model for the camera in order to decide probabilistically how the robot perceives each object. For each object that is perceived as matching the perceptual description *?pl* a local designator *?desig* is created and collected in the variable *desigs*. The last condition (*7*) estimates the time *?dt* after which the *look-for* behavior completes. Upon completion of the *look-for* behavior the projected interpretation process sends a success signal with the return value *?desigs* as its argument. The projected behavior consists of three events: two that change the world *begin(look-for(?pl, ?cam))* and *end(look-for(?pl, ?cam))*, which occurs *?dt* later. The third event changes the compu-

*project rule*  *look-for(?pl, ?cam)*
*(1)*    *if*    ( *loc(robot, ⟨?x,?y⟩)*
*(2)*         ∧ *normal-look-for-behavior(?cam, ?loc)*
*(3)*         ∧ *setof*  *?ob loc(?ob, ⟨?x,?y⟩) ?obs-here*
*(4)*         ∧ *sensor-model(?cam, ?sensor-model)*
*(5)*         ∧ *features(?pl,?features)*
*(6)*         ∧ *setof*  *?desig*
                     ( *member(?ob,?obs-here)*
                        ∧ *obj-seen(?ob, ?sensor-model)*
                        ∧ *perceived-properties(?ob, ?features, ?sensor-model, ?pl)*
                        ∧ *local-desig(?desig,?ob,?pl,?x,?y))*
                  *?desigs*
*(7)*         ∧ *look-time(⟨?x,?y⟩, ?features, ?dt))*
*(8)*    *with a delay of*  *0*  *occurs*  *mode transition*  *begin(look-for(?pl, ?cam))*
*(9)*    *with a delay of*  *?dt*  *occurs*  *mode transition*  *end(look-for(?pl, ?cam))*
*(10)*   *with a delay of*  *0*  *occurs*  *trigger fluent*  *visual-inputs-fluent(?cam)*
*(11)*   *with a delay of*  *0*  *occurs*  *set fluent*  *obs-pos-fluent ← ?seen)*
*(12)*   *with a delay of*  *0*  *occurs*  *succeed*  *?desigs*

Figure 15: A projection rule describing the behavior module *look-for*.

tational state of the structured reactive controller after passing *?dt* time units. This event pulses the fluent *visual-inputs-fluent(?cam)* and sets the fluent *obs-pos-fluent(?cam)*.

Besides asserting the events that take place during the execution of a plan we have to specify how these events change the world. This is done by using *effect rules*. One of them is shown in Figure! 16. The rule specifies that if at a time instant at which an event *end(look-for(?pl, ?cam))* occurs the state *visual-track(?desig, ?ob)* holds for some *?desig* and *?ob*, then the states *visual-track(?desig, ?ob)* will not (with a probability of *1.0*) persist after the event has occurred, i.e., they are clipped by the event.

*e→p rule*  VISUAL-TRACKING
*if* *visual-track(?desig, ?ob)*
*then*  *with probability*  *0.9*
     *event*  *end(look-for(?pl, ?cam))*
     *causes*  *clip*  *visual-track(?desig, ?ob))*

Figure 16: An *e→p* rule describing the effects of the event *end(look-for(?pl, ?cam))*.

## 5.5  Probabilistic Sampling-based Projection

So far we have looked at the issue of efficiently predicting an individual execution scenario. We will now investigate the issue of drawing inferences that are useful for planning based on sampled execution scenarios.

Recently, probabilistic sampling-based inference methods have been proposed to infer information from complex distributions quickly and with bounded risk (Fox, Burgard, Dellaert, & Thrun, 1999; Thrun, 2000). We will now discuss how we can use sampling-based projection for anticipating likely flaws with high probability.

Advantages of applying probabilistic sampling-based projection to the prediction of the effects of CRPs are that it works independently of the branching factor of the modes of the hybrid automaton and that it only constructs a small part of the complete PHAM.

But what kinds of prediction-based inferences can be drawn from samples of projected execution scenarios? The inference that we found most valuable for online revisions of robot plans is: do projected execution scenarios drawn from this distribution satisfy a given property $p$ with a probability greater than $\theta$? A robot action planner can use this type of inference to decide whether or not it should revise a plan to eliminate a particular kind of flaw: it should revise the plan if it believes that the flaw's likelihood exceeds some threshold and ignore it otherwise. Of course, such inferences can be drawn from samples only with a certain risk of being wrong. Suppose we want the planner to classify any flaw with probability greater than $\theta$ as to be eliminated and to ignore any flaw less likely than $\tau$. We assume that flaws with probability between $\tau$ and $\theta$ have no large impact on the robot's performance. How many execution scenarios should the plan revision module project in order to classify flaws correctly with a probability greater than 95%?

A main factor that determines the performance of sample-based predictive flaw detection is the *flaw detector*. A flaw detector classifies a flaw as to be eliminated if the probability of the flaw with respect to the robot's belief state is greater than a given threshold probability $\theta$. A flaw detector classifies a flaw as hallucinated if the probability of the flaw with respect to the robot's belief state is smaller than a given threshold $\tau$. So far we do not consider the severity of flaws, which is an obvious extension. Typically, we choose $\theta$ starting at 50% and $\tau$ smaller than 5%.

Specific flaw detectors can be realized that differ with respect to (1) the time resources they require; (2) the reliability with which they detect flaws that should be eliminated; and (3) the probability that they hallucinate flaws. That is, they signal a flaw that is so unlikely that eliminating the flaw would decrease the expected utility.

To be more precise consider a flaw $f$ that occurs in the distribution of execution scenarios of a given scheduled plan with respect to the agent's belief state with probability p. Further, let $X_i(f)$ represent the event that behavior flaw $f$ occurs in the $i$th execution scenario: $X_i(f) = 1$, if $f$ occurs in the $i$th projection and 0 otherwise.

The random variable $Y(f,n) = \sum_{i=1}^{n} X_i(f)$ represents the number of occurrences of the flaw $f$ in $n$ execution scenarios. Define a probable schedule flaw detector DET such that DET$(f,n,k)$ = *true iff* $Y(f,n) \geq k$, which means that the detector classifies a flaw $f$ as to be eliminated if and only if $f$ occurs in at least $k$ of $n$ randomly sampled execution scenarios. Thus DET$(f,n,k)$ works as follows. It first projects $n$ execution scenarios. Then it counts the number of occurrences of the flaw $f$ in the $n$ execution scenarios. If it is greater or equal to $k$ then the DET$(f,n,k)$ returns true, false otherwise.

Now that we have defined the schedule flaw detector, we can characterize it. Since the occurrence of schedule flaws in randomly sampled execution scenarios are independent from each other, the value of $Y(f)$ can be described by the binomial distribution $b(n,p)$. Using $b(n,p)$ we can compute the likelihood of overlooking a probable schedule flaw $f$ with probability $p$ in $n$ execution scenarios:

$$P(Y(f) < j) = \sum_{k=0}^{j-1} \left( \begin{array}{c} n \\ k \end{array} \right) * p^k * (1-p)^{n-k}$$

| | Prob. of Flaw $\theta$ | | | | |
|---|---|---|---|---|---|
| | 50% | 60% | 70% | 80% | 90% |
| DET(*f,3,2*) | 50.0 | 64.8 | 78.4 | 89.6 | 97.2 |
| DET(*f,4,2*) | 68.8 | 81.2 | 91.6 | 97.3 | 99.6 |
| DET(*f,5,2*) | 81.2 | 91.3 | 96.9 | 99.3 | 99.9 |

Figure 17: The table shows the probability of the flaw detectors DET(*f,i,2*) detecting flaws that have the probability $\theta$ = 50%, 60%, 70%, 80%, and 90%.

Figure 17 shows the probability that the flaw detector DET(*f,n,2*) for $n$ = 3,...,5 will detect a schedule flaw with probability $\theta$. The probability that the detectors classify flaws less likely than $\tau$ as to be eliminated is smaller than 2.3% (for all $n \leq 5$).

When using the prediction-based scheduling as a component in the controller of the robot office courier we typically use DET(*f,3,2*), DET(*f,4,2*), and DET(*f,5,2*) for the different experiments, which means a detected flaw is classified as probable if it occurs at least twice in three, four, or five detection readings.

Figure 18 shows the number of necessary projections to achieve $\beta = 95\%$ accuracy. For a detailed discussion see the work of Beetz et al. (1999).

| | $\theta$ | | | | | |
|---|---|---|---|---|---|---|
| | 1% | 10% | 20% | 40% | 60% | 80% |
| $\tau =.1\%$ | 1331 | 100 | 44 | 17 | 8 | 3 |
| $\tau =1\%$ | $\perp$ | 121 | 49 | 17 | 8 | 3 |
| $\tau =5\%$ | $\perp$ | 392 | 78 | 22 | 9 | 3 |

Figure 18: The table lists the number of randomly sampled projections needed to differentiate failures with an occurrence probability lower than $\tau$ from those that have a probability higher than $\theta$ with an accuracy of 95%.

The probabilistic sampling-based projection mechanism becomes extremely useful for improving robot plans during their execution once the execution scenarios can be sampled fast enough. At the moment a projection takes a couple of seconds. The overhead is mainly caused by recording the interpretation of RPL plans in a manner that is far too detailed for our purposes. Through a simplification of the models we expect an immediate speed up of up to one order of magnitude. It seems that with a projection frequency of about 100 Hz one could start tackling a number of realistic problems that occur at execution time continually.

## 6. Evaluation

We have validated our causal model of low-level navigation plans and their role in office delivery plans with respect to computational resources and qualitative prediction results in a series of experiments.

## 6.1 Generality

PHAMs are capable of predicting the behavior generated by flexible plans written in plan execution languages such as RAP (Firby, 1987) and PRS (Myers, 1996). To do so, we code the control structures provided by these languages as RPL macros. To the best of our knowledge PHAMs are the first realistic symbolic models of the sequencing layer of 3T architectures, the most commonly used software architectures for controlling intelligent autonomous robots (Bonasso et al., 1997). These architectures run planning and execution at different software layers and different time scales where a sequencing layer synchronizes between both layers. Each layer uses a different form of plan or behavior specification language. The planning layer typically uses a problem space plan, the execution layer employs feedback control routines that can be activated and deactivated. The intermediate layer typically uses a reactive plan language. The use of PHAMs enables 3T planning systems to make more realistic predictions of the robot behavior that is generated from their abstract plans. PHAMs are also capable of modeling different arbitration schemes and superpositions of the effects of concurrent control processes.

The causal models proposed here complement those introduced by Beetz (2000). He describes sophisticated models of object recognition and manipulation that allow for the prediction of plan failures including those that are caused by the robot overlooking or confusing objects, objects changing their location and appearance, and faulty operation of effectors. These models, however, were given for a simulated robot acting in a grid world. In this article, we have restricted ourselves to the prediction of behavior generated by modern autonomous robot controllers. Unfortunately, object recognition and manipulation skills of current autonomous service robots are not advanced enough for action planning. On the other hand, it is clear that action planning capabilities pay off much better if robots manipulate their environments and there is a risk of manipulating the wrong objects.

## 6.2 Assumptions and Restrictions

The control problem for autonomous robots is to generate effective and goal-directed control signals for the robot's perceptual and effector apparatus within a feedback loop. Plan-based robot control is a specialization of this control problem, in which the robot generates the control signals by maintaining and executing a plan that is effective and has a high expected utility with respect to the robot's dynamically changing belief state. This problem is so general that we cannot hope to solve it in this form.

In Computer Science it is common to characterize the computational problems a program can solve through the language in which the input for the program is specified. For example, we distinguish compilers for regular and context-free programming languages. The same is true for plan-based control of agents. Typically, planning problems are described in terms of an initial state description, a description of the actions available for the agents, their applicability conditions and effects, and a description of the goal state.

The three components of planning problems are typically expressed in some formal language. The problem solving power of the planning systems is characterized by the expressiveness of the languages for the three inputs. Some classes of planning problems are entirely formulated in propositional logic while others are formulated in first order logic. We further classify the planning problems with respect to the expressiveness of the action representations that they use; whether they allow for disjunctive preconditions, conditional effects, quantified effects, and model resource consumption. Some planning systems even solve planning problems that involve different kinds of uncertainty.

In contrast, SRCs use methods that make strong assumptions about plans to simplify the computational problems. As a consequence, SRCs can apply reliable and fast algorithms for the construction and installment of sub-plans, the diagnosis of plan failures, and for editing sub-plans during their execution. Making assumptions about plans is attractive because planning algorithms construct and revise the plans and can thereby enforce that the assumptions hold.

In a nutshell, the set of plans that an SRC generates is the reflexive, transitive closure of the routine plans with respect to the application of plan revision rules. Thus, to enforce that all plans have a property $Q$ it is sufficient that the routine plans satisfy $Q$ and that the revision rules preserve $Q$. These properties make it particularly easy to reason about the plans while the plans can still specify the same range of concurrent percept-driven behavior that RPL can. The properties of plans that play an important role in this article are their generality, flexibility, and reliability. These properties are achieved through careful design and hand-coding. As a consequence plan generation and revision can be performed by programmed heuristic rules. We believe, however, that such plans and rules can be learned from experience.

We make two other important assumptions. First, we assume that the tasks and the environment is benign and therefore behavior flaws do not result in disasters. This is important, because robots must make errors in order to learn the competent performance of tasks from experience. And only if the planner is allowed occasionally to propose worse plans we can apply fast planning methods based on Monte Carlo methods to improve the average performance of the robot.

Another design decision is that we do not explicitly represent the belief state of the robot, that is the probability distributions over the values of the state variables. This, however, does not need to imply that we cannot reason about inaccuracies and uncertainties of the robot's estimate of the world state. Beetz et al. (1998) describe how to couple plan-based high-level control with probabilistic state estimation. In this article the state estimator automatically computes and signals properties of the belief state such as the ambiguity and inaccuracy of state estimates to the plan-based controller. The plan-based controller, on the other hand, uses these signals in order to decide when to interrupt its missions to re-localize the robot.

## 6.3 Scaling Up

The causal models that we have described in Section 5 have been used for execution time planning for a robot office courier. The plans that have been projected were the original plans for this application and typically several hundreds of code lines long. The projected execution scenarios contained hundreds of events. Because the projection of single execution scenarios can cost up to a second, robots must revise plans based on very few samples. Thus, the robot can only detect probable flaws with high reliability.

The computational resources are mainly consumed by bookkeeping mechanisms that record the computational state of the robot at any time instant represented in the execution scenario and not by the mechanisms proposed in this article. The recorded computational state is used by the planning mechanisms in order to diagnose behavior flaws that are caused by discrepancies between the computational state of the robot and the state of the environment. The ability to reconstruct regularly updated fluent values is computationally very costly. We intend to provide programming constructs that let programmers declare the parts of the computational state that are irrelevant for planning and do not need to be recorded.

Even with this severe limitation we were able to show that with this preliminary implementation the robot can outperform controllers that lack predictive capabilities. The main source of inefficiency is the bookkeeping needed to reconstruct the entire computational state of the plan for any predicted time instant, an issue that we have not addressed in this article. Using a more parsimonious representation of the computational state we expect drastic performance gains.

### 6.4 Qualitatively Accurate Predictions

Projecting the plan listed in Figure 7 generates a timeline that is about 300 events long. Many of these events are generated through rescheduling the endogenous events (21 times). Figure 19 shows the predicted endogenous events (denoted by the numbered circles) and the behavior generated by the navigation plan in 50 runs using the robot simulator (we assume that the execution is interrupted in room A-111 because the robot realizes that the deadline can not be achieved). The qualitative predictions of behavior relevant for plan debugging are perfect. The projector predicts correctly that the robot will exploit the opportunity to go to location 5 while going from location 1 to 9.

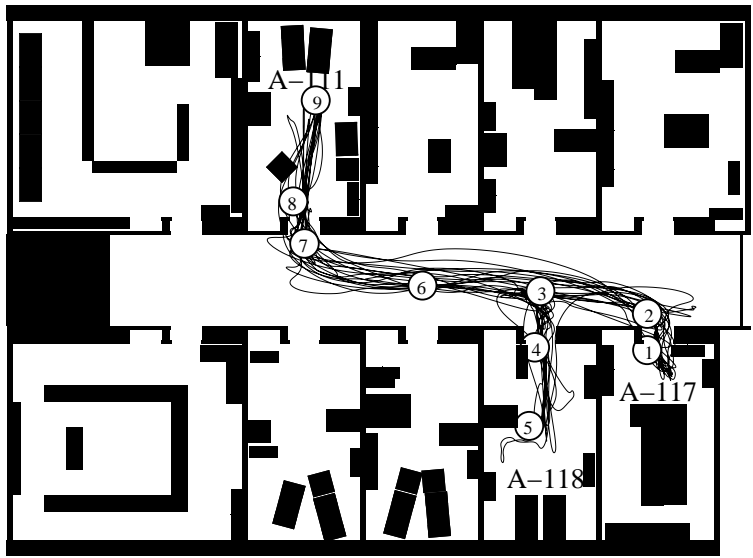

Figure 19: The figure shows the trajectories of multiple executions of the navigation plan and the events that are predicted by the symbolic plan projector.

### 6.5 Prediction-based Plan Debugging

Beetz (2002a and 2000) describes experiments showing that prediction-based plan debugging can improve the performance of robot controllers substantially.

## 7. Related Work

PHAMs represent external events, probabilistic action models, action models with rich temporal structure, concurrent interacting actions, and sensing actions in the domain of autonomous mobile robot control. There are many research efforts that formalize and analyze extended action representations and develop prediction and planning techniques for them. We know, however, only of approaches that address subsets of the aspects addressed by our representation. Related work comprises research on reasoning about action and change, probabilistic planning, numerical simulation, and qualitative reasoning.

**Reasoning about action and change.** Allen and Ferguson (1994) give an excellent and detailed discussion of important issues in the representation of temporally complex and concurrent actions and events. One important point that they make is that if actions have interfering effects then, in the worst case, causal models for all possible combinations of actions must be provided. In this paper, we have restricted ourselves to one kind of interference between actions: the transposition of movements which is the dominant kind of interference in physical robot behavior. In their article they do not address the issues of reasoning under uncertainty and efficiency with respect to computational resources.

A substantial amount of work has been done to extend the situation calculus (McCarthy, 1963) to deal with time and continuous change (Pinto, 1994; Grosskreutz & Lakemeyer, 2000a), exogenous (natural) actions (Reiter, 1996), complex robot actions (plans) (Levesque et al., 1997; Giacomo et al., 1997) using sensing to determine which action to execute next (Levesque, 1996; Lakemeyer, 1999) as well as probabilistic state descriptions and probabilistic action outcomes (Bacchus, Halpern, & Levesque, 1999; Grosskreutz & Lakemeyer, 2000b). The main difference to our work is that their representation is more limited with respect to the kinds of events and interactions between concurrent actions they allow. In particular, we know of no effort to integrate all of these aspects.

Some of the most advanced approaches in this area are formalizations of various variants of the high-level robot control language GOLOG, in particular CONGOLOG (Giacomo et al., 1997). Boutilier, Reiter, Soutchanski, and Thrun (2000) have applied decision theoretic means for optimally completing a partially specified GOLOG program. A key difference is that in the GOLOG approach the formalization includes the operation of the plan language whereas in our approach a procedural semantics realized through the high-level projector is used.

Hanks, Madigan, and Gavrin (1995) present a very interesting and expressive framework for representing probabilistic information, and exogenous and endogenous events for medical prediction problems. Because of their application domain they do not have to address issues of sophisticated percept-driven behavior as is done in this article.

**Extensions to Classical Action Planning Systems.** Planning algorithms, such as SNLP (McAllester & Rosenblitt, 1991), have been extended in various ways to handle more expressive action models and different kinds of uncertainty (about the initial state and the occurrence and outcome of events) (Kushmerick, Hanks, & Weld, 1995; Draper, Hanks, & Weld, 1994; Hanks, 1990). These planning algorithms compute bounds for the probabilities of plan outcomes and are computationally very expensive. In addition, decision-theoretic action planning systems (see Blythe, 1999, for a comprehensive overview) have been proposed in order to determine plans with the highest, or at least, sufficiently high expected utility (Haddawy & Rendell, 1990; Haddawy & Hanks, 1992;

Williamson & Hanks, 1994). These approaches abstract away from the rich temporal structure of events by assuming discrete atomic actions and ignore various kinds of uncertainty.

Planning with action models that have rich temporal structure has also been investigated intensively (Allen, Kautz, Pelavin, & Tenenberg, 1990; Dean, Firby, & Miller, 1988). IxTeT (Ghallab & Laruelle, 1994) is a planning system that has been applied to robot control and reasons about the temporal structure of plans to identify interferences between plan steps and resource conflicts. The planner/scheduler of the Remote Agent (Muscettola et al., 1998b) plans space maneuvers and experiments based on rich temporal causal models (Muscettola et al., 1998a; Pell et al., 1997). A good overview of the integration of action planning and scheduling technology can be found in an overview article by Smith, Frank, and Jonsson (2000). So far they have considered uncertainty only with respect to the durations of actions.

Kabanza, Barbeau, and St-Denis (1997) model actions and behaviors as state transition systems and synthesize control rules for reactive robots from these descriptions. Their approach can be used to generate plans that satisfy complex time, safety, and liveness constraints. These approaches too are limited with respect to the temporal structure of the (primitive) actions being modeled and the kinds of interferences between concurrent actions that can be considered.

**MDP-based planning approaches.** In recent years MDP (Markov decision process) planning has become a very active research field (Boutilier, Dean, & Hanks, 1998; Kaelbling, Cassandra, & Kurien, 1996). In the MDP approach robot behavior is modeled as a finite state automaton in which discrete actions cause stochastic state transitions. The robot is rewarded for reaching its goals quickly and reliably. A solution for such problems is a *policy*, a mapping from discretized robot states into, often fine-grained, actions.

MDPs form an attractive framework for action planning because they use a uniform mechanism for action selection and a parsimonious problem encoding. The action policies computed by MDPs aim at robustness and optimizing the average performance. A number of researchers have successfully considered navigation as an instance of Markov decision problems (MDPs) (Burgard et al., 2000; Kaelbling et al., 1996).

One of the main problems in the application of MDP planning techniques is to keep the problem encoding small enough so that the MDPs are still solvable. A number of techniques for complexity reduction can be found in the article written by Boutilier et al. (1998). Yet, it is still very difficult to solve big planning problems in the MDP framework unless the state and action spaces are well structured.

Besides reducing the complexity of specifying models for, and solving MDP problems, extending the expressiveness of MDP formalisms is a very active research area. Semi Markov decision problems (Bradtke & Duff, 1995; Sutton, Precup, & Singh, 1999) add a notion of continuous time to the discrete model of change used in MDPs: transitions from one state to another one no longer occur immediately, but according to a probability distribution. Others investigate mechanisms for hierarchically structuring MDPs (Parr & Russell, 1998), decomposing MDPs into loosely coupled sub-problems (Parr, 1998), and making them programmable (Andre & Russell, 2001). Rohani-manesh and Mahadevan (2001) propose an approach for extending MDP-based planning to concurrent temporally extended actions. All these efforts are steps towards the kind of functionality provided in the PHAM framework. Another relationship between the research reported here and the MDP research is that the navigation routines that are modeled with PHAMs are implemented on top of MDP navigation planning. Belker, Beetz, and Cremers (2002) use the MDP framework to learn action models for the improved execution of navigation plans.

The application of MDP based planning to reasoning about concurrent reactive plans is complicated by the fact that, in general, any activation and termination of a concurrent sub-plan might require a respective modification of the state and action space of the MDP.

Weaver (Blythe, 1995, 1996) is another probabilistic plan debugger capable of reasoning about exogenous events. Weaver uses Markov decision processes as its underlying model of planning. Weaver provides much of the expressiveness of PHAMs. Unlike Weaver, PHAMs are designed for reasoning about the physical behavior of autonomous mobile robots. Therefore, PHAMs add to Weaver's expressiveness in that they extensively support reasoning about concurrent reactive plans. For example, PHAMs can predict when the continuous effects of actions will trigger a concurrent monitoring process. PHAMs have built-in capabilities to infer the combined effects of two continuous motions of the robot.

**Qualitative reasoning about physical processes.**  Work in qualitative reasoning has researched issues in the quantization of continuous processes and focussed among other things on quantizations that are relevant to the kind of reasoning performed. Hendrix (1973) points out the limitations of discrete event representations and introduces a very limited notion of continuous process as a representation of change. He does not consider the influence of multiple processes on state variables. Hayes (1985) represents events as *histories*, spatially bounded, but temporally extended, pieces in time space, and proposes that histories which do not intersect do not interact. In Forbus' Qualitative Process Theory (Forbus, 1984) a technique called limit analysis is applied to predict qualitative state transitions caused by continuous events. Also, work on simulation often addresses the adequacy of causal models for a given range of prediction queries, an issue that is neglected in most models used for AI planning. Planners that predict qualitative state transitions caused by continuous events include EXCALIBUR (Drabble, 1993).

**Planning as model checking.**  Planning as model checking (Bertoli, Cimatti, & Roveri, 2001; Cimatti & Roveri, 2000) represents domains as finite-state systems. Planning problems are solved by searching through the state space, checking for the existence of a plan that satisfies the goals. Goals are formalized as logical requirements about the desired behavior for plans. Unlike planning as model checking we consider continuous control processes, plan interpretation as well as the physical effects of actions, and concurrency. This extended representational power comes at the cost of probably finding behavior flaws rather than proving their absence.

**Design and verification of embedded systems based on hybrid automata.**  The formalization of embedded software systems (Alur et al., 1997, 1996) using hybrid automata aims at proving critical aspects of the software rather than the physical effects of running this software. In our approach we have used the ideas of this research field as the basis of our conceptualization but added additional mechanisms to model the effects of actions and sensing mechanisms. Again, the additional complexity of our model is compensated by solving more restrictive inference problems: the detection of probable behavior flaws with high probability rather than safety of the system and the reachability of goals.

## 8. Conclusion

The successful application of AI planning to autonomous mobile robot control requires the planning systems to have more realistic models of the operation of modern robot control systems and the physical effects caused by their execution. In this article we have presented *probabilistic hybrid action models (*PHAM*s)*, which are capable of representing the temporal structure of continuous feedback control processes, their non-deterministic effects, several modes of their interferences, and exogenous events. We have shown that PHAMs allow for predictions that are, with high probability, qualitatively correct. We have also shown that powerful prediction-based inferences such as deciding whether a plan is likely to cause a flaw with a probability exceeding a given threshold can be drawn fast and with bounded risk.

We believe that equipping autonomous robot controllers with concurrent reactive plans and prediction-based online plan revision based on PHAMs is a promising way to improve the performance of autonomous service robots through AI planning both significantly and substantially.

The rules that we have used for projecting navigation behavior were hand-coded and plan and possibly even environment specific. On our research agenda is the development of transformational mechanisms for learning high performance and task specific plans. After having learned the plans the robot should then learn the projection rules by applying data mining techniques to the plan execution traces. To enable this approach we must invent novel representational mechanisms for the plans that allow for the automatic extraction of the rules. Initial steps into this direction can be found in the work of Belker et al. (2002), Beetz and Belker (2000), Beetz (2002b).

## References

Alami, R., Chatila, R., Fleury, S., Ingrand, M. H. F., Khatib, M., Morisset, B., Moutarlier, P., & Simeon, T. (2000). Around the lab in 40 days .... In *Proceedings of the IEEE International Conference on Robotics and Automation (ICRA 2000)*, pp. 88–94.

Allen, J., & Ferguson, G. (1994). Actions and events in interval temporal logic. *Journal of Logic and Computation*, *4*(5), 531–579.

Allen, J., Kautz, H., Pelavin, R., & Tenenberg, J. (Eds.). (1990). *Reasoning about Plans*. Morgan Kaufmann.

Alur, R., Henzinger, T., & Ho, P. (1996). Automatic symbolic verification of embedded systems. *IEEE Transactions on Software Engineering*, *22*(3), 181–201.

Alur, R., Henzinger, T., & Wong-Toi, H. (1997). Symbolic analysis of hybrid systems. In *Proceedings of the Thirtyssixth IEEE Conference on Decision and Control (CDC)*, pp. 702–707. IEEE Press.

Andre, D., & Russell, S. (2001). Programmable reinforcement learning agents. In *Advances in Neural Information Processing Systems 13, Papers from Neural Information Processing Systems (NIPS) 2000*, pp. 1019–1025. MIT Press.

Arkin, R. (1998). *Behavior based Robotics*. MIT Press.

Bacchus, F., Halpern, J., & Levesque, H. (1999). Reasoning about noisy sensors and effectors in the situation calculus. *Artificial Intelligence 111(1-2)*.

Beetz, M. (1999). Structured Reactive Controllers — a computational model of everyday activity. In Etzioni, O., Müller, J., & Bradshaw, J. (Eds.), *Proceedings of the Third International Conference on Autonomous Agents*, pp. 228–235.

Beetz, M. (2000). *Concurrent Reactive Plans: Anticipating and forestalling execution failures*, Vol. LNAI 1772 of *Lecture Notes in Artificial Intelligence*. Springer Publishers.

Beetz, M. (2001). Structured Reactive Controllers. *Journal of Autonomous Agents and Multi-Agent Systems. Special Issue: Best Papers of the International Conference on Autonomous Agents '99*, *4*, 25–55.

Beetz, M. (2002a). *Plan-based Control of Robotic Agents*, Vol. LNAI 2554 of *Lecture Notes in Artificial Intelligence*. Springer Publishers.

Beetz, M. (2002b). Plan representation for robotic agents. In *Proceedings of the Sixth International Conference on AI Planning and Scheduling*, pp. 223–232.

Beetz, M., Arbuckle, T., Bennewitz, M., Burgard, W., Cremers, A., Fox, D., Grosskreutz, H., Hähnel, D., & Schulz, D. (2001). Integrated plan-based control of autonomous service robots in human environments. *IEEE Intelligent Systems*, *16*(5), 56–65.

Beetz, M., Arbuckle, T., Cremers, A., & Mann, M. (1998). Transparent, flexible, and resource-adaptive image processing for autonomous service robots. In Prade, H. (Ed.), *Proceedings of the Thirteenth European Conference on Artificial Intelligence (ECAI-98)*, pp. 632–636.

Beetz, M., & Belker, T. (2000). Environment and task adaptation for robotic agents. In Horn, W. (Ed.), *Proceedings of the Fourteenth European Conference on Artificial Intelligence (ECAI-2000)*, pp. 648–652.

Beetz, M., Bennewitz, M., & Grosskreutz, H. (1999). Probabilistic, prediction-based schedule debugging for autonomous robot office couriers. In *Proceedings of the Twentythird German Conference on Artificial Intelligence (KI 99), Bonn, Germany*, pp. 243–254. Springer Publishers.

Beetz, M., Burgard, W., Fox, D., & Cremers, A. (1998). Integrating active localization into high-level control systems. *Robotics and Autonomous Systems*, *23*, 205–220.

Beetz, M., & Grosskreutz, H. (1998). Causal models of mobile service robot behavior. In Simmons, R., Veloso, M., & Smith, S. (Eds.), *Proceedings of the Fourth International Conference on AI Planning Systems*, pp. 163–170, Morgan Kaufmann.

Beetz, M., & Grosskreutz, H. (2000). Probabilistic hybrid action models for predicting concurrent percept-driven robot behavior. In *Proceedings of the Sixth International Conference on AI Planning Systems*, Toulouse, France. AAAI Press.

Beetz, M., & McDermott, D. (1992). Declarative goals in reactive plans. In Hendler, J. (Ed.), *Proceedings of the First International Conference on AI Planning Systems*, pp. 3–12, Morgan Kaufmann.

Beetz, M., & McDermott, D. (1996). Local planning of ongoing activities. In Drabble, B. (Ed.), *Proceedings of the Third International Conference on AI Planning Systems*, pp. 19–26, Morgan Kaufmann.

Beetz, M., & Peters, H. (1998). Structured reactive communication plans — integrating conversational actions into high-level robot control systems. In *Proceedings of the Twentysecond*

*German Conference on Artificial Intelligence (KI 98), Bremen, Germany*. Springer Publishers.

Belker, T., Beetz, M., & Cremers, A. (2002). Learning action models for the improved execution of navigation plans. *Robotics and Autonomous Systems*, *38*(3-4), 137–148.

Bertoli, P., Cimatti, A., & Roveri, M. (2001). Planning in nondeterministic domains under partial observability via symbolic model checking. In *Proceedings of the Seventeenth International Joint Conference on Artificial Intelligence (IJCAI-01)*. AAAI Press.

Blythe, J. (1995). AI planning in dynamic, uncertain domains. In *Extending Theories of Action: Formal Theory & Practical Applications: Papers from the 1995 AAAI Spring Symposium*, pp. 28–32. AAAI Press, Menlo Park, CA.

Blythe, J. (1996). Decompositions of Markov chains for reasoning about external change in planners. In Drabble, B. (Ed.), *Proceedings of the 3rd International Conference on Artificial Intelligence Planning Systems (AIPS-96)*, pp. 27–34. AAAI Press.

Blythe, J. (1999). Decision-theoretic planning. *AI Magazine*, *20*(2), 37–54.

Bonasso, P., Firby, J., Gat, E., Kortenkamp, D., Miller, D., & Slack, M. (1997). Experiences with an architecture for intelligent, reactive agents. *Journal of Experimental and Theoretical Artificial Intelligence*, *9*(1).

Boutilier, C., Dean, T., & Hanks, S. (1998). Decision theoretic planning: Structural assumptions and computational leverage. *Journal of Artificial Intelligence Research*, *11*, 1–94.

Boutilier, C., Reiter, R., Soutchanski, M., & Thrun, S. (2000). Decision-theoretic, high-level robot programming in the situation calculus. In *Proceedings of the Seventeenth AAAI National Conference on Artificial Intelligence*, pp. 355–362, Austin, TX.

Bradtke, S., & Duff, M. (1995). Reinforcement learning methods for continuous-time Markov decision problems. In Tesauro, G., Touretzky, D., & Leen, T. (Eds.), *Advances in Neural Information Processing Systems*, Vol. 7, pp. 393–400. MIT Press.

Brooks, R. (1986). A robust layered control system for a mobile robot. *IEEE Journal of Robotics and Automation*, *2*(1), 14–23.

Burgard, W., Cremers, A., Fox, D., Hähnel, D., Lakemeyer, G., Schulz, D., Steiner, W., & Thrun, S. (2000). Experiences with an interactive museum tour-guide robot. *Artificial Intelligence*, *114*(1-2), 3–55.

Cimatti, A., & Roveri, M. (2000). Conformant planning via symbolic model checking. *Journal of Artificial Intelligence Research (JAIR)*, *13*, 305–338.

Dean, T., Firby, J., & Miller, D. (1988). Hierarchical planning involving deadlines, travel time and resources. *Computational Intelligence*, *4*(4), 381–398.

Doherty, P., Granlund, G., Krzysztof, G., Sandewall, E., Nordberg, K., Skarman, E., & Wiklund, J. (2000). The WITAS unmanned aerial vehicle project. In *Proceedings of the Fourteenth European Conference on Artificial Intelligence (ECAI-00)*, pp. 747–755, Berlin, Germany.

Drabble, B. (1993). Excalibur: a program for planning and reasoning with processes. *Artificial Intelligence*, *62*, 1–40.

Draper, D., Hanks, S., & Weld, D. (1994). Probabilistic planning with information gathering and contingent execution. In *Proceedings of the Second International Conference on AI Planning Systems*, p. 31.

Firby, J. (1987). An investigation into reactive planning in complex domains. In *Proceedings of the Sixth National Conference on Artificial Intelligence*, pp. 202–206, Seattle, WA.

Forbus, K. (1984). Qualitative process theory. *Artificial Intelligence*, *24*, 85–168.

Fox, D., Burgard, W., Dellaert, F., & Thrun, S. (1999). Monte Carlo localization: Efficient position estimation for mobile robots. In *Proceedings of the Sixteenth National Conference on Artificial Intelligence*, Orlando, FL.

Ghallab, M., & Laruelle, H. (1994). Representation and control in IxTeT, a temporal planner. In Hammond, K. (Ed.), *Proceedings of the Second International Conference on AI Planning Systems*, pp. 61–67, Morgan Kaufmann.

Giacomo, G. D., Lesperance, Y., & Levesque, H. (1997). Reasoning about concurrent execution, prioritized interrupts, and exogene ous actions in the situation calculus. In *Proceedings of the Fifteenth International Joint Conference on Artificial Intelligence*, Nagoya, Japan.

Grosskreutz, H., & Lakemeyer, G. (2000a). cc-Golog: Towards more realistic logic-based robot controllers. In *Proceedings of the Seventeenth National Conference on Artificial Intelligence*.

Grosskreutz, H., & Lakemeyer, G. (2000b). Turning high-level plans into robot programs in uncertain domains. In *Proceedings of the Fourteenth European Conference on Artificial Intelligence (ECAI-00)*, pp. 548–552.

Haddaway, P., & Hanks, S. (1992). Representations for decision-theoretic planning: Utility functions for deadline goals. In Nebel, B., Rich, C., & Swartout, W. (Eds.), *Proceedings of the Third International Conference on Principles of Knowledge Representation and Reasoning*, pp. 71–82, Cambridge, MA. Morgan Kaufmann.

Haddawy, P., & Rendell, L. (1990). Planning and decision theory. *The Knowledge Engineering Review*, *5*, 15–33.

Hanks, S. (1990). Practical temporal projection. In *Proceedings of the Eighth National Conference on Artificial Intelligence (AAAI-90)*, pp. 158–163.

Hanks, S., Madigan, D., & Gavrin, J. (1995). Probabilistic temporal reasoning with endogenous change. In *Proceedings of the Eleventh Conference on Uncertainty in Artificial Intelligence*, pp. 245–254. Morgan Kaufmann.

Hayes, P. (1985). The second naive physics manifesto. In Hobbs, J. R., & Moore, R. C. (Eds.), *Formal Theories of the Commonsense World*, pp. 1–36. Ablex, Norwood, NJ.

Hendrix, G. (1973). Modeling simultaneous actions and continuous processes. *Artificial Intelligence*, *4*, 145–180.

Horswill, I. (1996). Integrated systems and naturalistic tasks. In: Strategic Directions in Computing Research, AI Working Group.

Kabanza, F., Barbeau, M., & St-Denis, R. (1997). Planning control rules for reactive agents. *Artificial Intelligence*, *95*, 67–113.

Kaelbling, L., Cassandra, A., & Kurien, J. (1996). Acting under uncertainty: Discrete Bayesian models for mobile-robot navigation. In *Proceedings of the IEEE/RSJ International Conference on Intelligent Robots and Systems*.

Konolige, K., Myers, K., Ruspini, E., & Saffiotti, A. (1997). The Saphira architecture: A design for autonomy. *Journal of Experimental and Theoretical Artificial Intelligence*, *9*(2).

Kushmerick, N., Hanks, S., & Weld, D. (1995). An algorithm for probabilistic planning. *Artificial Intelligence*, *76*, 239–286.

Lakemeyer, G. (1999). On sensing and off-line interpreting in golog. In Levesque, H., & Pirri, F. (Eds.), *Logical Foundations for Cognitive Agents*. Springer Publishers.

Levesque, H., Reiter, R., Lesperance, Y., Lin, F., & Scherl, R. (1997). Golog: A logic programming language for dynamic domains. *Journal of Logic Programming*, *31*, 59–84.

Levesque, H. J. (1996). What is planning in the presence of sensing. In *Proceedings of the Thirteenth National Conference on Artificial Intelligence*, pp. 1139–1146, Portland, OR.

McAllester, D., & Rosenblitt, D. (1991). Systematic nonlinear planning. In *Proceedings of the Ninth National Conference on Artificial Intelligence*, pp. 634–639, Anaheim, CA.

McCarthy, J. (1963). Situations, actions and causal laws. Tech. rep., Stanford University. Reprinted 1968 in Semantic Information Processing (M. Minsky ed.).

McDermott, D. (1991). A Reactive Plan Language. Research Report YALEU/DCS/RR-864, Yale University.

McDermott, D. (1992a). Robot planning. *AI Magazine*, *13*(2), 55–79.

McDermott, D. (1992b). Transformational planning of reactive behavior. Research Report YALEU/DCS/RR-941, Yale University.

McDermott, D. (1994). An algorithm for probabilistic, totally-ordered temporal projection. Research Report YALEU/DCS/RR-941, Yale University.

Muscettola, N., Morris, P., Pell, B., & Smith, B. (1998a). Issues in temporal reasoning for autonomous control systems. In Sycara, K., & Wooldridge, M. (Eds.), *Proceedings of the Second International Conference on Autonomous Agents (AGENTS-98)*, pp. 362–368. ACM Press.

Muscettola, N., Nayak, P., Pell, B., & Williams, B. (1998b). Remote Agent: to boldly go where no AI system has gone before. *Artificial Intelligence*, *103*(1–2), 5–47.

Myers, K. (1996). A procedural knowledge approach to task-level control. In Drabble, B. (Ed.), *Proceedings of the Third International Conference on AI Planning Systems*, pp. 158–165, Edinburgh, GB. AAAI Press.

Parr, R. (1998). Flexible decomposition algorithms for weakly coupled Markov decision problems. In Cooper, G. F., & Moral, S. (Eds.), *Proceedings of the Fourteenth Conference on Uncertainty in Artificial Intelligence (UAI-98)*, pp. 422–430, San Francisco. Morgan Kaufmann.

Parr, R., & Russell, S. (1998). Reinforcement learning with hierarchies of machines. In Jordan, M. I., Kearns, M. J., & Solla, S. A. (Eds.), *Advances in Neural Information Processing Systems*, Vol. 10. MIT Press.

Pell, B., Gat, E., Keesing, R., Muscettola, N., & Smith, B. (1997). Robust periodic planning and execution for autonomous spacecraft. In *Proceedings of the 15th International Joint Conference on Artificial Intelligence (IJCAI-97)*, pp. 1234–1239, San Francisco. Morgan Kaufmann.

Pinto, J. (1994). *Temporal Reasoning in the Situation Calculus*. Ph.D. thesis, Department of Computer Science, University of Toronto, Toronto, Ontario, Canada.

Reiter, R. (1996). Natural actions, concurrency and continuous time in the situation calculus. In *Proceedings of the Fifth International Conference on Principles of Knowledge Representation and Reasoning (KR-96)*, pp. 2–13.

Rohanimanesh, K., & Mahadevan, S. (2001). Decision-theoretic planning with concurrent temporally extended actions. In *Proceedings of the Seventeenth Conference on Uncertainty in Artificial Intelligence (UAI)*, pp. 472–479.

Schmitt, T., Hanek, R., Beetz, M., Buck, S., & Radig, B. (2002). Cooperative probabilistic state estimation for vision-based autonomous mobile robots. *IEEE Transactions on Robotics and Automation*, *18*(5), 670–684.

Simmons, R., Goodwin, R., Haigh, K., Koenig, S., & O'Sullivan, J. (1997). A modular architecture for office delivery robots. In *Proceedings of the First International Conference on Autonomous Agents*, pp. 245–252.

Smith, D., Frank, J., & Jonsson, A. (2000). Bridging the gap between planning and scheduling. *The Knowledge Engineering Review*, *15*(1), 47–83.

Sutton, R., Precup, D., & Singh, S. (1999). Between MDPs and semi-MDPs: A framework for temporal abstraction in reinforcement learning. *Artificial Intelligence 112(1-2)*, 181–211.

Thrun, S. (2000). Monte Carlo POMDPs. In *Advances in Neural Information Processing Systems 12*, pp. 1064–1070. MIT Press.

Thrun, S., Beetz, M., Bennewitz, M., Cremers, A., Dellaert, F., Fox, D., Hähnel, D., Rosenberg, C., Roy, N., Schulte, J., & Schulz, D. (2000). Probabilistic algorithms and the interactive museum tour-guide robot Minerva. *International Journal of Robotics Research*, *19*(11), 972–999.

Thrun, S., Bücken, A., Burgard, W., Fox, D., Fröhlinghaus, T., Hennig, D., Hofmann, T., Krell, M., & Schmidt, T. (1998). Map learning and high-speed navigation in RHINO. In Kortenkamp, D., Bonasso, R., & Murphy, R. (Eds.), *AI-based Mobile Robots: Case studies of successful robot systems*, pp. 21 – 52. MIT Press.

Williamson, M., & Hanks, S. (1994). Utility-directed planning. In *Proceedings of the Twelfth National Conference on Artificial Intelligence*, p. 1498, Seattle, WA.